\documentclass{JASSS}
\usepackage{multirow}
\usepackage{makecell}
\usepackage{subcaption}
\usepackage{placeins}
\usepackage{float}
\usepackage{url}

\title{Towards Credible Agent-Based Policy Simulations: Disentangling Opportunities and Preferences in a Financial Inclusion Case Study of Egypt}

\reviewcopy{false} 

\author[1,2]{Alba Aguilera}
\author[3]{Georgina Curto}
\author[1]{Nardine Osman}
\author[2]{Ahmed Al-Awah}

\affil[1]{Artificial Intelligence Research Institute (IIIA-CSIC)}
\affil[2]{United Nations Economic and Social Commission for Western Asia (UN-ESCWA)}
\affil[3]{United Nations University Institute in Macau (UNU Macau)}

\email{aaguilera@iiia.csic.es}

\usepackage{natbib}
	\setcitestyle{authoryear,round,aysep={}}
\usepackage{appendix}
	
\begin{document}
\maketitle 



\begin{abstract}
Credibility is a central topic for agent-based models intended to support policy-making. Simulations must not only represent the target scenarios and their core dynamics but also demonstrate that their assumptions, parameters, and outputs are empirically grounded and sufficiently accurate for their intended use. This paper addresses this challenge by presenting a general modelling framework, aligned with the Capability Approach, for building credible policy simulations that rely on data and domain-expert knowledge. It then demonstrates how it can be contextualised and implemented to study the social challenge of financial inclusion in Egypt. The framework builds from an agent-based model that, in this particular context, represents heterogeneous individuals and firms behaving according to their financial states, barriers, opportunities, and preferences. The model is fitted to real-world data in two stages, initialisation and calibration, which respectively build representative synthetic populations and estimate behavioural parameters. By fixing the feasibility parameters, which determine agents' opportunities, and calibrating preference parameters across different population groups, we are able to distinguish and analyse the role of institutional and social barriers in the system, as well as the role of agents' motivations and priorities. This calibration stage provides transparent and group-specific hypotheses about the drivers of observed financial-inclusion gaps, which can further be analysed as gaps between agents' opportunities and realised outcomes, a very relevant insight for policy-making. This paper is thus a step towards improving the credibility and usefulness of policy simulations, strengthening the relationship between the model, the real target system, and the stakeholders who will use it. The code is available at: \url{https://www.comses.net/codebase-release/df8383cb-f49b-4f09-8cce-73603b59adcc/}. 
\end{abstract}


\parano{}



\section{Introduction}

Policy analysis and design is a complex task in which microdata and aggregate indicators alone rarely provide sufficient insight into how different interventions play out across a diverse and context-dependent population. Agent-based models (ABMs) are well suited to policy analysis because they are able to represent systems from the bottom up: individual agents are assigned diverse attributes, constraints, and behavioural parameters, and aggregate outcomes emerge from their simulated behaviour and interactions~\citep{bonabeau2002agent, macal2010tutorial, epstein1999generative}. Policies can then be implemented as top-down changes in their distribution of resources or existing barriers and opportunities, and later evaluated through simulated trajectories and outcomes. This makes ABMs especially useful when policy questions concern not only whether aggregated indicators, such as income, increase, but also for whom, through which mechanisms, and under which contextual assumptions~\citep{hammond2015policy, gilbert2018policy, lempert2002policy}. 

However, ABMs also face a major credibility challenge: their usefulness depends on whether their assumptions, parameters, and simulated outcomes are empirically grounded, transparent and valid for the stakeholders in question. Calibration and validation are central to this challenge. Calibration refers to the process of determining parameters so that the model can reproduce relevant empirical patterns or theoretically expected behaviour, while validation assesses whether the model is adequate for its intended purpose and whether its outputs can be considered credible in relation to the real-world phenomenon being represented. Both processes remain difficult in ABMs because models' designs are highly \textit{ad hoc}, their outputs are usually stochastic, parameter spaces can be very large, and many relevant behavioural mechanisms depend on highly contextual and latent perceptions, norms, or institutional constraints that are only partially observed in available data~\citep{windrum2007empirical,fagiolo2007critical,guerini2017validation,grazzini2017bayesian,platt2020comparison,mcculloch2022uncertainty}. Therefore, strengthening the relationship between the model, the real target system, and the people who know the system (domain-experts) in order to contextualize, simplify and contrast the simulation's design and outcomes is central to model's credibility. 

To address this, the main contribution of this paper is a general framework for building policy simulations that are simultaneously aligned with human development principles and empirically grounded in both data and domain-expert knowledge. 
Our framework is grounded in the Capability Approach (CA), a widely recognised theory of human development that shifts attention from resources people hold, or the outcomes they happen to reach, to the real opportunities (the capabilities) available to people given their circumstances~\citep{sen1999, nussbaum2011, robyens2017}. This shift is what makes the CA valuable for policy analysis: an intervention never changes outcomes directly, but expands or narrows people's opportunities through changes in the institutional rules or resources that constrain them. Evaluating a policy by an aggregate outcome alone can therefore mislead, since the aggregate does not reveal why a group falls short of a desired outcome. People may lack the opportunity to reach it, held back by personal barriers repeatedly identified as obstacles to the SDGs~\citep{escwa_annual_sdg_review_2025,escwa_social_development_report_5}; or they may simply not prefer it, given contextual differences in their priorities. The CA addresses these two relevant questions for policy design: whether people \textit{can} act, and whether they \textit{choose} to, and this distinction shapes both how our agents decide and how we evaluate what they do. The CA also emphasises that all of its elements are context-dependent, which is why the presented framework constantly draws on both data and domain-expert knowledge.

The specific context, design, calibration techniques, and validation strategy of a simulation will differ from one application to another, but the designed stages of our CA-aligned agent-based modelling framework remain: (1) a mapping that connects the simulation's elements to those of the CA; (2) an initialisation stage that builds the agent population and its environment from micro- and macro-level data; (3) a calibration stage that estimates agents' preferences using likelihood techniques and then interprets their results; and (4) a validation stage that, among other steps yet to be implemented, submits the calibrated preferences in the form of a list of hypotheses to domain-expert review. Together, these stages aim to ensure holistic, interpretable, and empirically grounded policy simulations that generate credible insights. 

We demonstrate the framework through a case study for financial inclusion in Egypt. Financial inclusion, defined as access to and use of formal financial services, is a core pillar for inclusive growth and poverty reduction, and it has become a priority across the Middle East and North Africa (MENA) region. Yet, persistent gaps remain in account ownership, digitally enabled accounts, and the effective use of services such as savings, credit, and digital payments, with especially pronounced disparities for women, youth, small and medium enterprises (SMEs), lower-income households, and rural populations~\citep{escwa2025financialInclusiveness, worldbank_global_findex_2025}. These gaps do not only depend on resources and supply (e.g., proximity of providers and available infrastructure) but also depend on context-specific constraints and perceptions, such as documentation requirements, digital and financial literacy, trust in institutions, or religious concerns. Several institutionally verified reports argue that such complex policy challenges require granular data, disaggregated indicators, and advanced analytical tools to design targeted interventions for different population groups~\citep{escwa_ispar_fii_2025,escwa_annual_sdg_review_2025,escwa_social_development_report_5}, which is precisely what the simulation presented here aims to provide.

To analyse these broad financial inclusion gaps, the agent-based model represents individuals and firms as agents, with heterogeneous financial states and constraints or barriers that shape their transitions towards different outcomes, such as formal account ownership, digital payments usage, formal saving and borrowing, or firm formalisation. Using simulation-based Bayesian calibration, we estimate the preference parameters given a fixed configuration of feasibility or opportunity constraints. We then identify each calibrated preference movement with respect to the prior expected value and interpret it as the behavioural component of choice once feasibility constraints have been accounted for. On this basis, the policy simulation aims to provide candidate interpretations of the calibrated behaviour, to be handed to and validated by a panel of domain experts, rather than treated as settled by the model, and ultimately, a group-specific analysis of the financial inclusion gap. By evaluating the model through the CA, and defining a set of financially-related capabilities, we aim to deliver disaggregated and interpretable policy insights framed in terms of opportunities, not just resources and outcomes. 

The remainder of the paper is organised as follows. We first situate our contribution within the literature on agent-based policy modelling, calibration, and validation. We then present the framework methodology through the financial-inclusion simulation, covering the CA-mapping, initialisation, and calibration stages, as well as the designed domain-expert validation. Next, we discuss the model's numerical and conceptual limitations. Finally, we conclude by outlining future work, including the later stages of the validation plan.

\section{Related Work}
\label{sec: relwork}

The usefulness of an ABM for policy making case studies highly depends on the credibility and transparency of the simulation's mechanisms and results. This has motivated a broad body of literature on calibration, estimation, validation and verification of ABMs, especially in economics, epidemiology, ecology, and urban or social simulation~\citep{windrum2007empirical,fagiolo2007critical,guerini2017validation,mcculloch2022uncertainty,belfrage2024simulating,belfrage2024credible}.

Within social simulation, a recent line of work combines the CA with ABMs~\citep{aguilera2025agent,aguilera2026}, modeling agent behavior with Markov decision processes and reinforcement learning to learn optimal behavioral strategies~\citep{aguilera2025agent,aguilera2026}. This behaviour is calibrated against synthetic data using Bayesian known likelihood techniques. This paper instead adopts \emph{stochastic behavioural rules}: rather than optimising a reward, each agent samples an action from a distribution shaped by the feasibility of the action and its calibrated preference. These stochastic rules are a deliberate baseline: they are simple, interpretable, transparent to stakeholders, and designed to be made more complex only if the application requires it, by being replaced with (or benchmarked against) more structured decision models such as explicit MDPs, reinforcement learning, or LLM-based choice. On this baseline we extend this CA-aligned line in three directions relevant to policy simulation credibility: (i) far higher scalability, both in the size of the agent population and in the range of policies that can be implemented; (ii) richer, non-binary feasibilities and preferences built and calibrated from real-world data; and (iii) a validation plan that subjects both the fixed feasibility assumptions and the calibrated preference movements to domain expert review. To situate these contributions, we next review model credibility, which heavily relies on calibration and validation techniques. 

A first stream of literature focuses on formal calibration and estimation of ABMs' parameters. ABM calibration remains difficult because simulations are often stochastic, computationally expensive, high-dimensional, and characterized by intractable likelihoods or equifinality, where multiple parameter combinations reproduce similar aggregate patterns~\citep{mcculloch2022uncertainty,dyer2024blackbox}. Most existing work calibrates parameters by explicitly comparing simulated outputs with observed data~\citep{windrum2007empirical,fagiolo2007critical}, using approaches such as simulated minimum distance~\citep{grazzini2015estimation} or Bayesian estimation~\citep{grazzini2017bayesian}. \citet{platt2020comparison} compares several calibration strategies and finds Bayesian estimation very competitive in benchmark economic ABMs~\citep{grazzini2017bayesian,platt2020comparison}. This has motivated a broader set of likelihood-free and approximate methods, including ABC, surrogate modelling, particle MCMC, black-box Bayesian inference, and uncertainty-quantification pipelines~\citep{lamperti2018surrogates,lux2022particlemcmc,dyer2024blackbox,mcculloch2022uncertainty}. In our case, we explicitly evaluate a binomial observation likelihood for empirical targets using probabilities estimated from repeated ABM simulations. Therefore, our approach can be best understood as simulation-based Bayesian calibration with a constructed binomial likelihood. 

A second stream of research concerns empirical validity of the model, i.e. whether ABMs reproduce and explain relevant observed patterns~\citep{barde2017empirical,guerini2017validation,mcculloch2022uncertainty}. Rather than being a single statistical test, validation is usually understood as the process of assessing whether a model is sufficiently adequate for its intended purpose. In ABMs, this assessment is especially challenging because every model is substantially different in its micro-level assumptions about who the agents are, its own rules for how they act and interact between them and the environment, and its own emergent aggregate outputs. Because these ingredients differ from one model to the next, there is no universal and standard validation approach across the interdisciplinary field of agent-based modelling and simulation. Several authors therefore argue that validation should be treated as a structured set of evidence about the relation among the model, the target system, and the stakeholders who will use the model~\citep{windrum2007empirical,fagiolo2007critical,guerini2017validation,collins2024validation}. These would include, for example, checks that the model's mechanisms are plausible to those who know the system, that it reproduces observed patterns in data at several levels, or that it behaves sensibly under known interventions.

\begin{figure*}[t]
    \centering
    \includegraphics[width=0.99\linewidth]{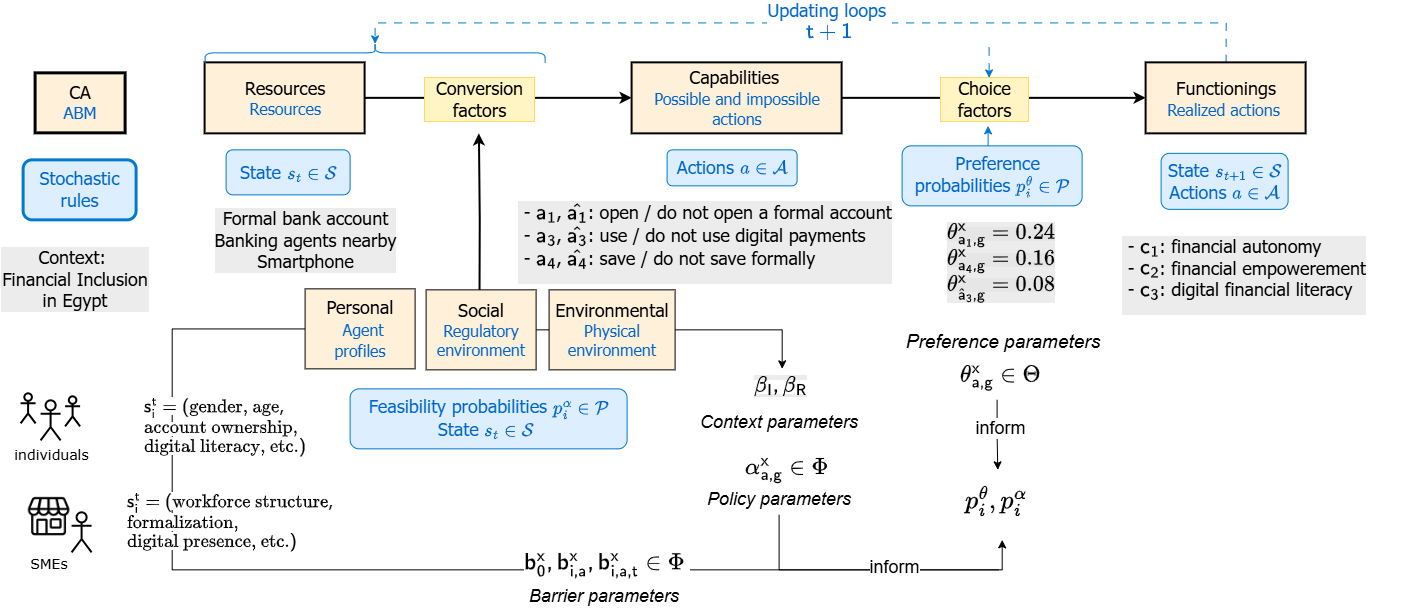}
\caption{CA-aligned simulation framework from~\citep{aguilera2025agent,aguilera2026}, operationalised for the financial inclusion context in Egypt. }
    \label{fig:CA}
\end{figure*}

For policy-oriented ABMs, credibility also depends on traceability, documentation, stakeholder communication, and a clear statement of the model's intended use~\citep{belfrage2024credible,belfrage2024simulating}. Certain streams of work address situations in which data-driven calibration cannot, by itself, settle whether a model's assumptions or estimated parameters are adequate, and turns instead to structured judgement from those familiar with the target system. This idea is called \emph{face validity}: presenting a model's assumptions and behaviour to people knowledgeable about the real system and asking whether they are plausible, as one stage within a broader validation process that also covers input--output validation~\citep{naylor1967verification, morgan2014use}. Drawing on this literature, we use established guidance to determine the composition of the expert panel~\citep{sutherland2015policy}, the model components and results to be reviewed~\citep{grimm2005pattern}, and the structure of the elicitation procedure~\citep{hemming2018practical} in order to assess the plausibility of the model.

\section{Methodology through Implementation}
\label{sec: method}
This section presents the framework and its implementation together: each stage is stated in general terms and then instantiated for the financial-inclusion case study in Egypt. We begin with the \emph{CA mapping}\ref{subsec: mapping}, which defines who the agents are, how do they make decisions, and how do we evaluate them. Then, the \emph{initialisation} stage\ref{subsec: init} builds the agent population and its environment from micro- and macro-level data, the \emph{calibration} stage\ref{subsec:calibration} estimates the preference parameters against observed targets given a fixed feasibility configuration and interprets the results, and the \emph{validation} stage\ref{subsec:expertvalidation} submits these interpretations to domain-expert review.

Validation is not completed in this paper, but framed as a multi-stage plan. We design and describe its first stage, a structured domain-expert review of the model's feasibility assumptions and calibrated preferences, and leave its execution to future work, together with the later stages: historical (input--output) validation against observed pre- and post-policy outcomes, and benchmarking against alternative decision-support tools.


\subsection{Capability Approach Mapping}
\label{subsec: mapping}

Building from the CA implies distinguishing between \textit{feasibility} and \textit{preference} in the agents' decision-making. For instance, in the context of financial inclusion, an agent may not own a formal account for two different reasons: the action is not feasible for him or her (e.g. because account fees are too high), or the action is feasible but not chosen by the agent (e.g. because of a lack of trust in banks). These two components define the main behavioural parameters influencing agents' choices in the CA-aligned simulation framework from~\citep{aguilera2025agent,aguilera2026}, shown in Fig.~\ref{fig:CA}. The figure follows the canonical CA elements (in yellow) from resources to functionings, and shows how each of its elements is computationally operationalised in the model (in blue) and contextualised with the specific financial inclusion elements (in grey). We explain each of these elements below:

\begin{itemize}
    \item \textbf{Resources} are the commodities, assets and services available to an agent in a given context (e.g. bank account, smartphone, assets usable as collateral, bank branches and agents nearby, etc.). They are represented within the agent's state $s_t\in\mathcal{S}$.
    \item \textbf{Conversion factors} are the personal, social and environmental conditions that determine how those resources can be turned into real opportunities. In our case they comprise the agent's personal profile (e.g. gender, age, urbanicity), including their reported barriers (e.g. income level, distance to banks, financial literacy), the social and regulatory environment (e.g. account fees, documentation strictness, lending guarantees), and the physical and digital environment (e.g. agent and infrastructure density of bank branches, ATMs, microfinance institutions, digital lenders, fintech firms, etc.). The environment information is summarised within the context parameters $\beta_R$ and $\beta_I$ defined in the initialisation section\ref{subsec: init}. This information is either stored in the agent's state $s_t\in\mathcal{S}$ or represented through probabilities $p \in \mathcal{P}$ that influence the feasibility of actions, i.e. how possible the actions are for the agent. 
    \item \textbf{Capabilities} are \textit{what agents are able to do and be}, given their resources and conversion factors, represented in the model by the actions $a\in \mathcal{A}$. We use the word (deprived) \textit{opportunity} further on in the paper to refer to capabilities that are (not) sufficiently feasible for the agent, except in the evaluation, where we will rely on aggregated, financially-related \textit{capabilities}.  
    \item \textbf{Choice factors} are the behavioural motivators that lead an agent to prioritise one available action over another. They are represented by a set of disaggregated parameters $\theta \in \Theta$ across heterogeneous agent types and are meant to be calibrated against observed data. 
    \item \textbf{Functionings} are \textit{what agents end up doing and being}. They are represented as the actions the agent actually realises, which deterministically update its state to $s_{t+1}$ and are recorded through the outcome indicators. We use the word \textit{outcomes} further on in the paper. 
\end{itemize}

In practical terms, the CA framework influences both the design of the agents' decision-making and the evaluation of simulation outcomes. On the \textit{decision-making} side, we separate whether an action is feasible for an agent from whether the agent prefers that action, so that a low observed outcome can either be attributed to very restrictive constraints or to behavourial tendencies towards not performing an action. On the \textit{evaluation} side, we separate real opportunities available to an agent from the outcomes the agent actually realises, so that a shortfall in outcomes can be traced to lack of opportunities, particularly among population subgroups identified as opportunity-deprived. 

In this implementation, agents behave according to stochastic rules informed by their feasibilities and preferences. Each simulation step represents one month, a time scale that reasonably reflects the frequency with which individuals and firms make financial decisions. At every step, each agent chooses an action based on whether the action is feasible and aligned with their preferences, and the realized actions update the agent's state and thereby the population aggregates over time. The ultimate goal is to analyse the resulting model aggregates to study hypothetical policy-making scenarios. 
The following subsections describe (i) the agents' profile and environment\ref{subsec:data}, (ii) the agents' decision-making mechanism\ref{subsec: actions}, and (iii) the evaluation of their opportunities and outcomes\ref{subsec: capabilities}. 

\begin{figure}
    \centering
    \includegraphics[width=0.55\textwidth]{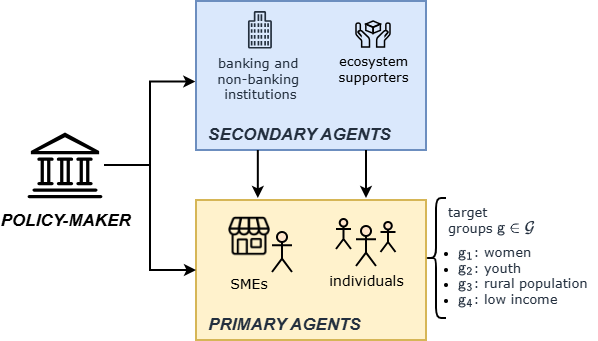}
    \caption{ Main agents involved in the financial inclusion context for the Arab region, defined according to domain expert knowledge. }
    \label{fig:agents1}
\end{figure}

\subsubsection{Agent population and Environment}
\label{subsec:data}

As shown in Fig.~\ref{fig:agents1}, we distinguish between primary agents, secondary agents and policy-makers. Primary agents are the main actors in the simulation and, in this case study, comprise two agent types $x$: individuals and firms. In particular, we focus on small and medium enterprises (SMEs), instead of general firms. Each type is disaggregated into policy-relevant target groups $g \in \mathcal{G}$, where $\mathcal{G}$ denotes the set of groups along which we disaggregate and analyse the simulation's opportunities and outcomes: women, youth, rural populations, and low-income individuals; and women-led, rural-managed, and low-capital firms. Groups are not exclusive, so an agent may belong to several at once, which is what allows the model to represent intersectional differences. 

Secondary agents form part of the environment in which primary agents operate. They comprise banking and non-banking institutions, such as banks or microbanks, fintechs, incubators or accelerators; and ecosystem supporters such as non-profits and multi-lateral organisations. They are not simulated as decision-making agents: their role enters the model through the context parameters $\beta_I$, $\beta_R$ that raise or lower the feasibility of primary agents' actions. Policy-makers, such as central banks and government agencies, are not considered agents, but rather prospective users of the tool. Their interventions influence the primary and secondary agents, either through direct changes in the feasibility of actions or through the context parameters, which in turn affect feasibility. The initialisation section\ref{subsec: init} describes how each of these elements is derived from data for the financial-inclusion case study in Egypt.

\begin{table*}[t]
\centering
\small
\caption{Central opportunity--action--outcome mapping used in the simulation.}
\label{tab:actions}
\begin{tabular}{p{1.0cm}p{4.0cm}p{6.6cm}p{3.0cm}}
\toprule
\textbf{Agent} & \textbf{Central Capabilities $c \in \mathcal{C}$} & \textbf{Actions $a \rightarrow c$} & \textbf{Indicators $i \rightarrow c$} \\
\midrule
\multirow{4}{*}{\centering Indiv.} & \textbf{$c_1$: Financial autonomy}: freedom to independently own and operate financial instruments. & \makecell[l]{$a_1$, $\hat{a}_1$ open / do not open a formal account  } & Traditional/digital account ownership rate. \\
& \textbf{$c_2$: Financial empowerment}: freedom to access credit and borrow from reliable sources for productive growth. & \makecell[l]{$a_4$, $\hat{a}_4$ save / do not save formally  \\ $a_5$, $\hat{a}_5$ borrow / do not borrow formally  \\ $a_6$, $\hat{a}_6$ borrow / do not borrow informally} & Formal/informal borrowing rate. \\
& \textbf{$c_3$: Digital financial literacy}: ability to use digital financial services effectively. & \makecell[l]{$a_2$, $\hat{a}_2$ adopt / do not adopt mobile money \\ $a_3$, $\hat{a}_3$ use / do not use digital payments}& Digital/traditional payments and remittances. NGO reached users. \\
\midrule
\multirow{4}{*}{\centering SMEs} & \textbf{$c_5$: Formalization}: freedom to operate as a recognized enterprise. & \makecell[l]{$a_7$, $\hat{a}_7$ obtain / do not obtain a business license \\ $a_8$, $\hat{a}_8$ obtain / do not obtain a commercial registration \\ $a_9$, $\hat{a}_9$ adopt / do not adopt accounting books} & Share of formalized SMEs; share of SMEs with valid license. \\
& \textbf{$c_6$: Digital financial literacy}: ability to adopt digital channels for business transactions. & \makecell[l]{$a_{10}$, $\hat{a}_{10}$ digitize / do not digitize sales/payments } & Share of SMEs with digitized sales/payments. \\
& \textbf{$c_7$: Employment quality}: freedom to create stable and protected jobs. & \makecell[l]{$a_{11}$, $\hat{a}_{11}$ hire / do not hire non-household workers \\ $a_{12}$, $\hat{a}_{12}$ insure / do not insure workers} & Share of employer SMEs with insured workers. \\
\bottomrule
\end{tabular}
\end{table*}

\subsubsection{Feasibility, Preference and Choice}
\label{subsec: actions}

Let $\mathcal{A}^x$ denote the set of modelled actions for each primary agent type $x$, defined in Table~\ref{tab:actions}. For each action $a \in \mathcal{A}^x$, we define a complementary action $\hat{a} \in \mathcal{A}^x$. The action $a$ represents a transition toward a financial inclusion outcome, such as opening a formal account; whereas $\hat{a}$ represents continuation in the current state, such as remaining without one. For simplicity we write the equations for a generic action $a$, with the understanding that $a$ may denote either a transition action or its complement. 

To distinguish \textit{feasibility} from \textit{preference} within a stochastic rule-based decision architecture, we operationalise two of the CA elements introduced above as explicit parameter sets. In particular, \textit{conversion factors} determine whether an action is feasible for an agent $i$ at time $t$, and are captured by the feasibility set
\[
\Phi =
\{b_{0,a},\; b_{i,a,t},\; b_{a,t},\; \alpha^{x}_{a,t},\; \alpha^{x}_{a,g,t}
\mid a \in \mathcal{A}^x,\ g \in \mathcal{G}\},
\]
while \textit{choice factors} determine which action is preferred by the agent, and are captured by the preference set
\[
\Theta =
\{\theta^x_a, \theta^x_{a,g}
\mid a \in \mathcal{A}^x,\ g \in \mathcal{G}\},
\]

Both sets are indexed by agent type $x$ and action $a$, with a general term and a group-specific term $g \in \mathcal{G}$, so that the model can represent intersectional differences in constraints and choices across opportunities and groups. The two sets enter the behavioural rule at different points: $\Phi$ determines what an agent \emph{can} do, and $\Theta$ determines what it \emph{wants} to do. Within $\Phi$, the $b$ terms are held fixed to describe the constraints agents face due to their personal and environmental circumstances, while the $\alpha$ terms are adjustable multipliers, neutrally initialised at $1$, used as the channel through which a hypothetical policy widens or narrows those constraints. On the other hand, preference parameters are calibrated against target data and used to describe how desirable one action is relative to another, once feasibility constraints have been considered.

For each agent $i$, the model computes the feasibility probability to describe how feasible it is to perform action $a$ at time $t$, considering the agent's conversion factors:
\begin{equation}
p^{\alpha}_{i,a,t}
=
b_{0,a}
\cdot
b_{i,a,t}
\cdot b_{a,t}
\cdot \alpha^{x}_{a,t}
\cdot \prod_{g \in \mathcal{G}_i} \alpha^{x}_{a,g,t}, 
\label{eq:feasibility}
\end{equation}
where $\mathcal{G}_i \subseteq \mathcal{G}$ denotes the set of target groups to which agent $i$ belongs. The baseline term $b_{0,a}$ captures the assumed starting feasibility of action $a$ before agent-specific barriers are applied. $b_{i,a,t}$ captures individual- or firm-level perceived barriers derived from survey data, while $b_{a,t}$ captures contextual constraints derived from macro-level data and represented through the context parameters $\beta$. For each action, $b_{a,t}$ is the product of the context multipliers relevant to that action. All numerical values are specified in Tables~\ref{tab:feasibility_assumptions} and~\ref{tab:feasibility_assumptions2}, justified according to domain-expert knowledge in the reports. 

To test hypothetical interventions, we consider the policy multipliers. The general policy multipliers $\alpha^{x}_{a,t}$ capture interventions that affect all agents equally, while group-specific $\alpha^{x}_{a,g,t}$ parameters capture interventions targeted at particular groups. For example, an expansion of digital-payment infrastructure may increase digital feasibility for all agents, while digital financial literacy campaigns targeted at women or rural populations would primarily affect these corresponding groups. We acknowledge that, in general terms, policies could be affecting agents' preferences too. However, in this model prototype, we prioritize the effect of policies on the agents' opportunities. In this way, preference parameters can be calibrated and fixed to capture people's motivations, which are more stable over the years, and derivable from data, not like their opportunities. 

The multiplicative nature of the probability equation, Eq.~(\ref{eq:feasibility}), treats each term as an independent gate that scales feasibility down from its baseline. Therefore, an agent facing several simultaneous constraints is less likely to act than one facing a single constraint, producing group-specific and intersectional opportunity effects.
Additionally, the feasibility of some actions further depends on the realisation of other actions. For instance, formal saving ($a_4$) and formal borrowing ($a_5$) require a formal financial account. If an individual does not yet have a formal account, their feasibility is set to zero. After account opening ($a_1$) is realized, these actions can become feasible in later steps, subject to the remaining barriers and context terms. Other actions do not fully require a previous action, but become easier after one has been realized. For example, digital-payment use ($a_3$) is easier when the agent already has an account, mobile wallet, or digital account. For SMEs, formalization ($a_7$-$a_9$) and employment actions ($a_{11}$-$a_{12}$) also depend on firm state: registration, accounting books, and worker insurance become easier after earlier formalization or hiring steps.

\begin{figure*}[t]
    \centering
    \begin{subfigure}{0.92\linewidth}
        \centering
        \includegraphics[width=\linewidth]{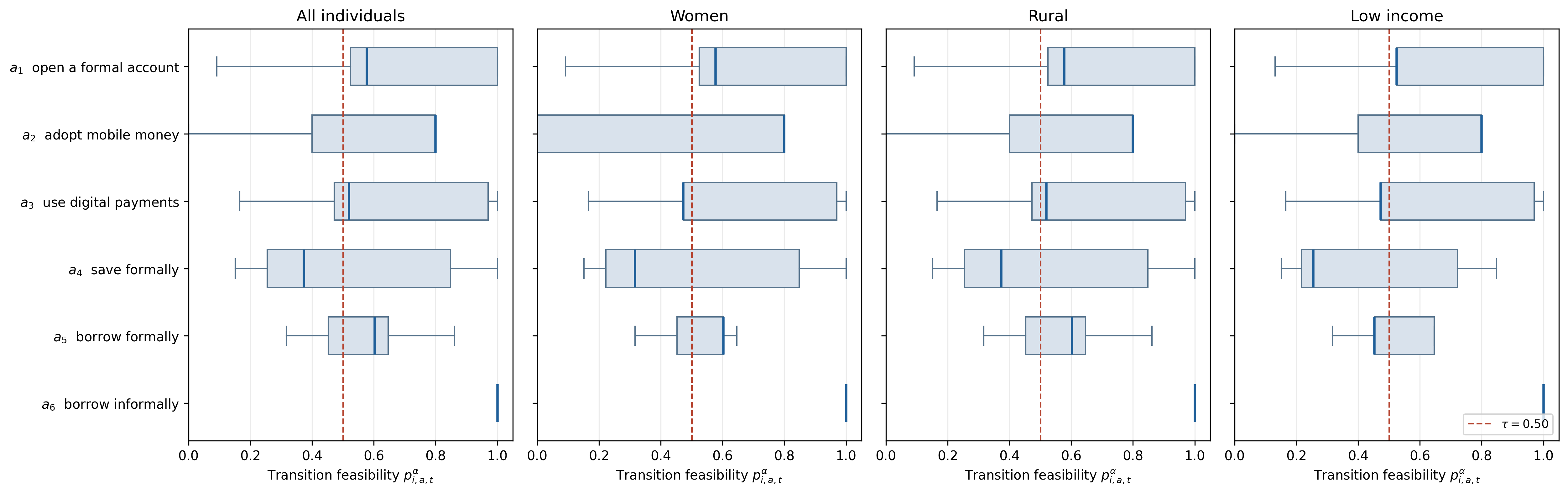}
        \caption{Individuals}
        \label{fig:feasibility_indiv}
    \end{subfigure}

    \vspace{1.5ex}
    \begin{subfigure}{0.92\linewidth}
        \centering
        \includegraphics[width=\linewidth]{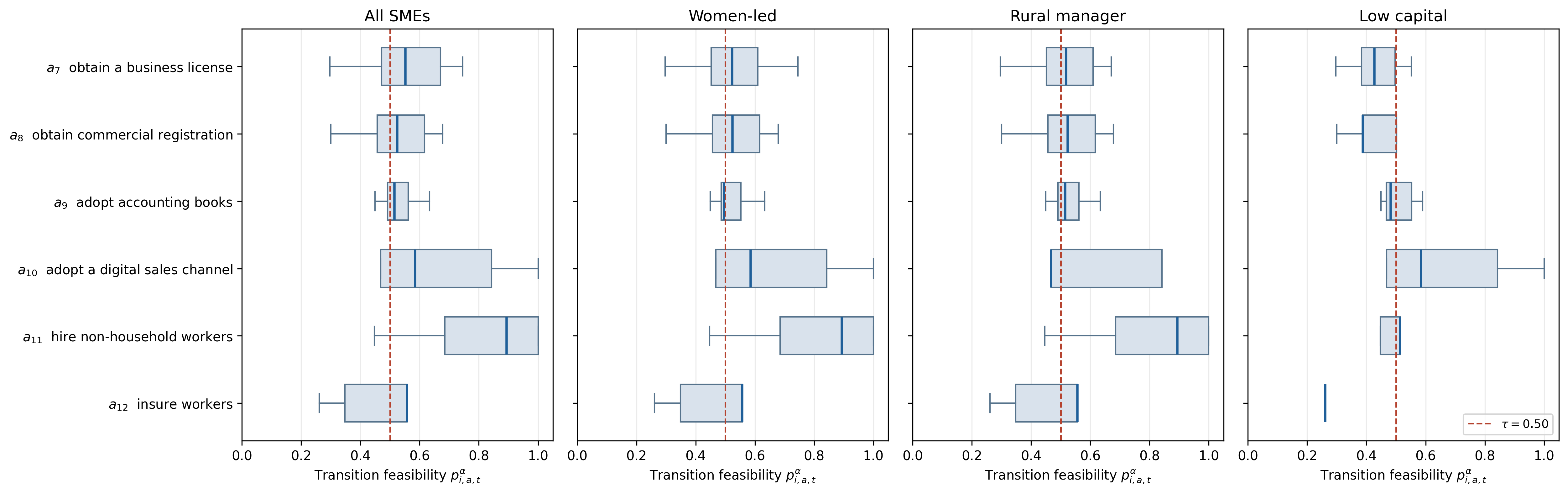}
        \caption{Small enterprises}
        \label{fig:feasibility_sme}
    \end{subfigure}
    \caption{Feasibility probability $p^{\alpha}_{i,a,t}$ for each action at initialization, disaggregated by target group, for (a) individuals and (b) small enterprises. The dashed line marks the opportunity threshold $\tau=0.5$.}
    \label{fig:feasibility}
\end{figure*}

Once feasibility probabilities have been computed, agents choose among the actions based on their preferences. In our model, agents are more likely to select actions that are both feasible and aligned with their preferences. Based on this assumption, for each action $a \in \mathcal{A}^{x}$, the model computes the behavioural rule 
\begin{equation}
r_{i,a, t}
= 
p^{\alpha}_{i,a,t}
\cdot
\theta^{x}_{a}
\cdot
\prod_{g \in G_i \cap \mathcal{G}_a}
\theta^{x}_{a,g}
\cdot
\lambda_a ,
\label{eq:score}
\end{equation}
which functions as an action score that increases when an action is more feasible, more preferred, or both. Here, $p^{\alpha}_{i,a,t}$ is the feasibility probability of action $a$, $\theta^{x}_{a}$ is the general preference multiplier for action $a$, and $\theta^{x}_{a,g}$ are group-specific preference multipliers. The factor $\lambda_a$ is a transition scaling factor (fixed at $0.24$ to set the timescale over which the calibrated preferences reproduce the observed outcome levels). Without it, an action that is both feasible and preferred would be selected by most agents within the first few steps, and the population aggregates would saturate almost immediately. This is unrealistic: opening an account or registering a business is a step an agent takes once, at some point over a period of years, not one that most agents take at once. We therefore apply $\lambda_a$ to transition actions only, leaving complements unscaled, so that in any given step an agent is more likely to remain in its current state than to move out of it, and realised transitions accumulate gradually over the simulated horizon.

This action score is then normalized over the action set to obtain the preference probability
\begin{equation}
{p}^{\theta}_{i,a,t}
= 
\frac{r_{i,a,t}}
{\sum_{a' \in \mathcal{A}^{x}} r_{i,a',t}}, 
\label{eq:averagepreference}
\end{equation}

from which the agent samples one action each timestep. To reflect real-world behaviour through this rule, preference parameters are initialized and calibrated to microdata reflecting their behaviour and outcomes. In our case study, we use financial usage information from individual- and firm-level surveys (Table~\ref{tab:preference_prior_assumptions} in Annex A). 

\subsubsection{Evaluation: Opportunities and Outcomes}
\label{subsec: capabilities}

Building from the CA, we evaluate the simulation through opportunities and outcomes. An action is considered a meaningful opportunity for an agent when it is sufficiently feasible. The opportunity set of agent $i$ at time $t$ is defined as 
\begin{equation}
\mathcal{O}_{i,t}
=
\{a \in \mathcal{A}^x : p^{\alpha}_{i,a,t} \geq \tau \}, 
\label{eq:opportunityset}
\end{equation}

where $\tau$ is a feasibility threshold fixed at $0.5$, so that an action counts as available when an agent is at least as likely to be able to perform it as not. The threshold is a modelling choice rather than an empirical quantity, and Fig.~\ref{fig:feasibility} shows where it falls relative to the distribution of feasibility probabilities for each action and target group at the initial time step.

After the agent samples an action from the probability distribution in Eq.~\ref{eq:averagepreference}, if the selected action belongs to the opportunity set $\mathcal{O}_{i,t}$, the corresponding state is updated deterministically, following the feedback loop in Fig.~\ref{fig:CA}. For example, if the selected action is $a_1$, opening a formal account, the state variable indicating formal account ownership is updated to one. The feasibility probabilities of other actions are then recomputed, since realizing one action may further expand the agent's opportunity set (e.g. realized $a_1$ may unlock $a_3$ and $a_4$), as emphasised by the CA. In contrast, if the selected action is not in the opportunity set, the action is recorded as desired but not feasible (what we refer to as a \textit{deprived opportunity}), and the agent state remains unchanged. For example, an agent may select $a_5$, formal borrowing, because the preference for formal credit is high, but if the feasibility probability is very low, the loan may not be obtained. Thus, realizing an action may either produce a state transition or leave the agent's state unchanged. We are interested in evaluating these changes, especially when hypothetical policy interventions are implemented in the simulation. 

To evaluate them, however, we do not rely only on individual opportunities, especially if the set of modelled actions becomes large. Following~\citet{nussbaum2011}, who introduces the notion of \textit{central capabilities} as a list of essential opportunities of such importance that a society should secure them to at least a threshold level, we group these opportunities into a small set of domain-specific central capabilities $\mathcal{C}$, such as financial autonomy, empowerement or literacy  (Table~\ref{tab:actions}). We do not adopt Nussbaum's universal list, which addresses human dignity in general, but rather define a set specific to financial inclusion, grounded in the financial-capability frameworks used by the World Bank, OECD and CGAP~\citep{kempson2013measuring,oecd2022toolkit,mazer2018financialcapability}. Each action is then linked to a central capability through a mapping $\mathcal{R} \subseteq \mathcal{A} \times \mathcal{C}$ together with an associated outcome indicator. Table~\ref{tab:actions} thus defines the evaluative space of the simulation: the set of central opportunities $\mathcal{C}$ and outcome indicators. From here on, "opportunity" and "outcome" denote aggregate scores, and we speak of an individual action's opportunity only when comparing at the action level.

Let $\mathcal{A}_c = \{a : (a,c)\in\mathcal{R}\}$ denote the actions contributing to central opportunity $c$, and let $\omega_a$ be an evaluation weight capturing the positive (restoring) or negative (depriving) contribution of action $a$ to $c$. The agent's \textit{opportunities} and \textit{outcomes} are the weighted shares of contributing actions that are, respectively, available and realised:
\begin{equation}
\mathrm{opportunity}_{i,c,t}=\frac{\sum_{a\in\mathcal{A}_c}\omega_a\,o_{i,a,t}}{\sum_{a\in\mathcal{A}_c}|\omega_a|},
\qquad
\mathrm{outcome}_{i,c,t}=\frac{\sum_{a\in\mathcal{A}_c}\omega_a\,y_{i,a,t}}{\sum_{a\in\mathcal{A}_c}|\omega_a|}.
\label{eq:capability}
\end{equation}

where $y_{i,a,t}\in\{0,1\}$ indicates whether agent $i$ has realised action $a$, and $o_{i,a,t}=\max\!\big(\mathbf{1}[\,p^{\alpha}_{i,a,t}\ge\tau\,],\,y_{i,a,t}\big)$ indicates whether $a$ is a genuine opportunity (Eq.~\ref{eq:opportunityset}). The weights $\omega_a$ are set from domain-expert criteria, and in this prototype we use equal, positive weights ($\omega_a=1$), so that every contributing action weighs equally in its central opportunity. Because $o_{i,a,t}\ge y_{i,a,t}$, the outcome is always contained in the opportunity, and their difference is the \textit{opportunity gap}
\begin{equation}
\mathrm{gap}_{i,c,t} = \mathrm{opportunity}_{i,c,t} - \mathrm{outcome}_{i,c,t},
\label{eq:capabilitygap}
\end{equation}
which captures the share of opportunities that are not converted into realised outcomes, averaged across the agents in a target group. A large gap indicates that opportunities exist but are not being converted into outcomes, whereas jointly low opportunity and outcome indicate structural deprivation. This is what makes the CA-aligned evaluation useful for policy analysis: it lets you analyse interventions that are known to expand opportunities way before affecting aggregate outcomes, or even without affecting them at all. For example, reducing documentation requirements for opening a bank account may immediately increase the feasibility of opening an account, while account ownership may respond more slowly or not respond at all because it depends on other behavioral or contextual constraints, such as people's financial literacy. 
\begin{table*}[t!]
    \centering
    \small
    \setlength{\tabcolsep}{4pt}
    \caption{Micro- and macro-level data sources used for initialisation.}
    \label{tab:datasources}
    \begin{tabular}{p{0.055\textwidth} p{0.165\textwidth} p{0.125\textwidth} p{0.09\textwidth} p{0.48\textwidth}}
        \toprule
        \textbf{Level} & \textbf{Dataset} & \textbf{Size (cases $\times$ columns)} & \textbf{Geographic granularity} & \textbf{Information} \\
        \midrule
        \multirow{3}{*}{\centering Micro} & $ds_1$:  Global Findex Survey ~\citep{worldbank_global_findex_2025} & 1001 $\times$ 178 & Nation & \textbf{Individual-level socio-demographic characteristics and financial inclusion outcomes}: account ownership, saving, borrowing, payments, insurance, digital access and use, and self-reported barriers to formal financial service use. \\
        & $ds_2$: Egypt Labour Market Panel Survey~\citep{erf_elmps_2023} & 70{,}000 $\times$ 4{,}500 (IND) & Village & \textbf{Individual- and household-level socio-economic characteristics, labour-market status}, and, where available, \textbf{enterprise and finance-related modules} for synthetic population enrichment. \\
        & $ds_3$: Egypt Labor Force Survey~\citep{oamdi2025_hlfs_lfs_2024_v1} & 297{,}000 $\times$ 103 (IND) & Region & \textbf{Labour-market characteristics} used to structure the synthetic population, including employment status, sector, occupation, and income-related variables. \\
        \midrule
        \multirow{4}{*}{\centering Macro} & \multicolumn{3}{p{0.38\textwidth}}{$ds_4$: Financial Access Survey~\citep{imf_fas_2025}} & \textbf{Infrastructure and financial service providers:} counts of commercial banks, microfinance institutions, credit unions and cooperatives, and insurance corporations. \\
        & \multicolumn{3}{p{0.38\textwidth}}{$ds_5$: Indicators~\citep{itida_venture} and reports~\citep{researchandmarkets_egypt_bnpl_2026update}} & \textbf{Non-bank financial inclusion ecosystem:} venture capital and accelerator counts, typical ticket size, fintech focus, cohort size, and graduation rate. \\
        & \multicolumn{3}{p{0.38\textwidth}}{$ds_6$: Indicators~\citep{fra_non_banking} and reports~\citep{entlaq_fintech_egypt_2024,entrepreneurme_investing_egypt_startup_2024}} & \textbf{Digital and non-bank finance:} BNPL providers, active users, market size, digital lenders, average loan size, and default rate. \\
        & \multicolumn{3}{p{0.38\textwidth}}{$ds_7$: Indicators~\citep{cbe_fi_indicators_2025} and reports~\citep{oecd_sme_egypt_2025,rise_financial_inclusion_egypt_2024}} & \textbf{Institutional and support ecosystem:} central bank and related institutional sources used to characterise payment systems, financial inclusion infrastructure, and support networks for MSMEs and digital finance. \\
        \bottomrule
    \end{tabular}
\end{table*}

\section{Initialisation}
\label{subsec: init}
At the initial time step, the simulation should provide an approximate representation of the real-world system under study. The initialisation procedure ensures this by representing the agent population and its surrounding environment using empirical and contextual data. 
We use the micro- and macro-level datasets, together with reports from institutionally verified sources, listed in Table~\ref{tab:datasources}, to generate:
\begin{itemize}
    \item Synthetic populations of varying sizes, using a bayesian network model like the one specified in~\citep{aguilera2025population}, and individual and firm-level survey microdata in Table~\ref{tab:datasources}. We sample two synthetic populations for each primary agent type: 
    \begin{itemize}
        \item Synthetic population of individuals, representing people over 16 years old. Each data row contains socio-demographic information (e.g. gender, age, education, income quintile, workforce status, rural/urban residence, etc.), labour information (e.g. main/secondary employment status, contract, insurance, wage, hours, etc.), and several financial-inclusion variables such as financial ownership and access (e.g. account at a financial institution, mobile money account, digitally enabled account, self-efficacy proxy), financial history and usage (e.g. has formal account, uses digital payments, saves or borrows formally) and perceived barriers (e.g. lack of money, distance to services, high fees). 
        \item Synthetic population of SMEs, representing household non-agricultural enterprises. These are owned, managed, or operated by household members, and each enterprise is linked, when possible, to a synthetic individual acting as its manager. This population therefore mainly captures small enterprises, including both informal and formal businesses. Each data row contains firm characteristics (age, activity, workforce structure, insurances, licenses, etc.), and financial information (e.g. annual earnings, taxes, digital usage, formality). 
    \end{itemize}
    We validate the resulting synthetic populations by comparing them against the source microdata. After analysing marginal and joint distributions across the datasets, we commit to a $20,000$-person synthetic population, associated with $4,318$ synthetic firms for this prototype. 
    \item A set of context parameters $\beta$, split into infrastructure ($\beta^I$) and regulatory ($\beta^R$) groups, representing the agents' regulatory and infrastructure environment. These act as environmental conversion factors within the term $b_{a,t}$ of Eq.~(\ref{eq:feasibility}). They are derived from aggregated macrodata and reports in Table~\ref{tab:datasources}, and further specified in Annex~\ref{annex:assumptions} for this implementation. 
    \begin{itemize}
        \item  Infrastructure context parameters ($\beta^I$) represent the role of the secondary agents, including banking and non-banking institutions in the physical financial infrastructure (e.g. number of bank branches, ATMs, microfinance institutions, cooperatives) and digital financial infrastructure (e.g. number of fintech firms, payment services, mobile banking, digital lenders, BNPL); and also ecosystem supporters (e.g. BNPL providers, venture-capital funds, accelerators, NGOs, MSME support infrastructures). We normalise these counts into two density factors, $\beta_I$ (traditional) and $\beta_D$ (digital), which scale down the feasibility of access-dependent actions from its baseline. For example, account opening is scaled by $0.50+0.50\,\beta_I$: with no traditional infrastructure ($\beta_I=0$) its feasibility is halved but not eliminated (an account can still be opened through minimal or occasional channels) whereas full infrastructure ($\beta_I=1$) leaves it at its baseline. A policy that expands infrastructure raises feasibility by moving these factors up toward one.
        \item Regulatory context parameters ($\beta^R$) represent the institutional conditions affecting the difficulty of accessing or using formal financial services, including account fees, documentation strictness, lending guarantees. They are instantiated as qualitative levels (low / medium / high) and mapped to normalised factors $\beta_f,\beta_d,\beta_T\in[0,1]$ (e.g.\ $0.3/0.6/0.9$) that scale actions feasibility. For example, account opening is scaled by $(1-0.30\,\beta_f)(1-0.25\,\beta_d)$: a higher fee level ($\beta_f$) or stricter documentation ($\beta_d$) each lowers feasibility proportionally, with fees weighted slightly more ($0.30$ vs $0.25$) because they are the most frequently reported barrier to account ownership. A policy that eases this regulatory conditions raises feasibility by moving these factors up towards one. 
    \end{itemize}

\end{itemize}
\section{Calibration}
\label{subsec:calibration}

Calibration involves adjusting model parameters so that simulated outcomes reproduce observed patterns in the data. In this model only the preference set $\Theta$ is calibrated, while the feasibility set $\Phi$ is held fixed. Because feasibility is fixed, every posterior movement in $\Theta$ is, by construction, an estimate of \emph{preference}: the behavioural component of choice that reproduces the observed data conditional on the assumed opportunity structure. We therefore interpret each estimated posterior value relative to its prior as a hypothesis of agents' behaviour, which is exactly what the expert panel is asked to check in the validation stage\ref{subsec:expertvalidation}). 

The policy parameters remain adjustable for policy-makers to test hypothetical policy scenarios. During these analyses, the calibrated preferences would be held fixed, reflecting the assumption that behavioural preferences remain approximately stable over the time horizon considered, while policies primarily affect the opportunities of agents. Differences in simulated outcomes would therefore be interpreted as the consequences of changes in the opportunity structure given the underlying behavioural preferences assumed.

\subsection{Bayesian Calibration Procedure}

Following Bayesian approaches to agent-based model calibration~\citep{grazzini2017bayesian,platt2020comparison}, calibration is implemented as a Bayesian likelihood-based procedure, following the equation
\begin{equation}
p(\Theta \mid y^{obs})
\propto
\mathcal{L}(y^{obs}\mid \Theta)\,p(\Theta),
\label{eq:bayes_calibration}
\end{equation}
where $p(\Theta \mid y^{obs})$ is the posterior distribution of the preference parameters after observing the vector of observed calibration targets $y^{obs}$, $\mathcal{L}(y^{obs}\mid \Theta)$ is the likelihood of observing these targets under preference parameters $\Theta$, and $p(\Theta)$ is the prior distribution over preference parameters. The procedure consists of defining prior distributions over behavioural parameters, simulating the model under candidate parameter values, evaluating the likelihood of the observed targets given the simulated outcomes, and sampling from the resulting posterior distribution, which we do using a Metropolis-Hastings algorithm~\citep{hastings1970monte}.

According to the decision rule in Eq.~\ref{eq:score}, preference parameters are positive multiplicative factors, with $\theta_j=1$ representing a neutral preference. To facilitate Bayesian estimation while preserving this positivity constraint, each parameter is estimated on the logarithmic scale. For every preference multiplier $\theta_j \in \Theta$, where $j$ indexes either a general $(a)$ or a group-specific term $(a,g)$, we define the log-transformed $z_j = \log(\theta_j)$.

The transformed parameters are assigned Gaussian priors $z_j \sim \mathcal{N}(\mu_j,\sigma_j^2)$, where the prior mean $\mu_j$ represents the expected preference before calibration and the prior variance $\sigma_j^2$ controls how strongly the calibration is allowed to deviate from that expectation. Rather than centring all priors neutrally at $\theta_j=1$, we use the observed shares in the data to initialise the priors for the preference parameters associated with complement and subgroup-specific actions, i.e.,  $\theta_{\hat{a}}$ and $\theta_{a,g}$. In this way, we give the calibration a reasonable starting direction for preference parameters, with the main goal of reflecting heterogeneous subgroup patterns in the data in the posterior calibrated parameters. For each of these parameters, we compute a data-informed prior based on the observed ratio
\[
R_j =
\frac{share^{obs}_{num,j}}{share^{obs}_{ref,j}},
\]
where the numerator and reference depend on the parameter being calibrated. For complement parameters $\theta_{\hat{a}}$, the ratio compares observed non-adoption and adoption shares to indicate how much this action is not realized with respect to how much it is. For subgroup parameters $\theta_{a,g}$, the ratio compares subgroup and overall adoption shares to indicate how much the action is realized by a subgroup with respect to a general agent. To prevent extreme observed ratios from dominating calibration, the ratios are shrunk toward the neutral value before defining the prior means. These ratios, and resulting prior centers $\theta^0_j = \sqrt{R_j}$ and $\mu_j=\log(\theta^0_j)$, are reported in Table~\ref{tab:preference_prior_assumptions}.

\subsection{Targets and Likelihood}
Let $\mathcal{M}$ denote the set of calibration targets extracted from the microdata listed in Table~\ref{tab:datasources}. Each target $m \in \mathcal{M}$ has an observed value $y_m^{obs}$ and a corresponding simulated statistic
$\widehat{y}_m(\Theta \mid \Phi,\beta,\lambda)$,
computed after running the simulation for a fixed number of steps. 
Since all targets are empirical proportions, such as the share of individuals with a formal account or the share of SMEs with a business license, we evaluate them using a binomial likelihood. For each target $m$, the observed proportion $y_m^{obs}$ is converted into an observed count $k_m = n_m^{eff}y_m^{obs}$, where $n_m^{eff}$ denotes the effective sample size. The log-likelihood contribution of target $m$ is
\[
\ell_m(\Theta)
=
w_m
\left[
k_m \log(\widehat{y}_m)
+
(n_m^{eff}-k_m)\log(1-\widehat{y}_m)
\right],
\]
where $\widehat{y}_m(\Theta \mid \Phi,\beta,\lambda)$ is the simulated proportion and $w_m$ is a target-specific weight, giving information about how much target $m$ should matter relative to other targets in the posterior objective. It is set to $1$ for individual targets, $0.6$ for SME targets and $0.55$ for subgroup targets. The posterior objective is therefore
\begin{equation}
\log p(\Theta \mid y^{obs})
= \sum_{m \in \mathcal{M}} \ell_m(\Theta)
+ \sum_{\theta_j \in \Theta} \log p(z_j).
\end{equation}
The first term rewards preference values that reproduce the observed calibration targets. The second term is the log-prior contribution, rewarding preference values that do not deviate a lot from the prior ones. The calibration accepts large movements in preference parameters only when they improve the fit to observed targets enough to compensate for this prior penalty.

\subsection{Results and Parameter Plausibility}
To test our proposed model, calibration is carried out with $20{,}000$ agents and $20$ simulation steps, repeated across $3$ random seeds with $2$ Monte Carlo runs per likelihood evaluation. The likelihood includes $m=59$ calibration targets ($12$ aggregate targets and $47$ subgroup targets), covering every modelled action for both individuals and SMEs, so that the calibration fits aggregate and subgroup outcomes.

\begin{figure*}[t]
    \centering
    \includegraphics[width=0.99\linewidth]{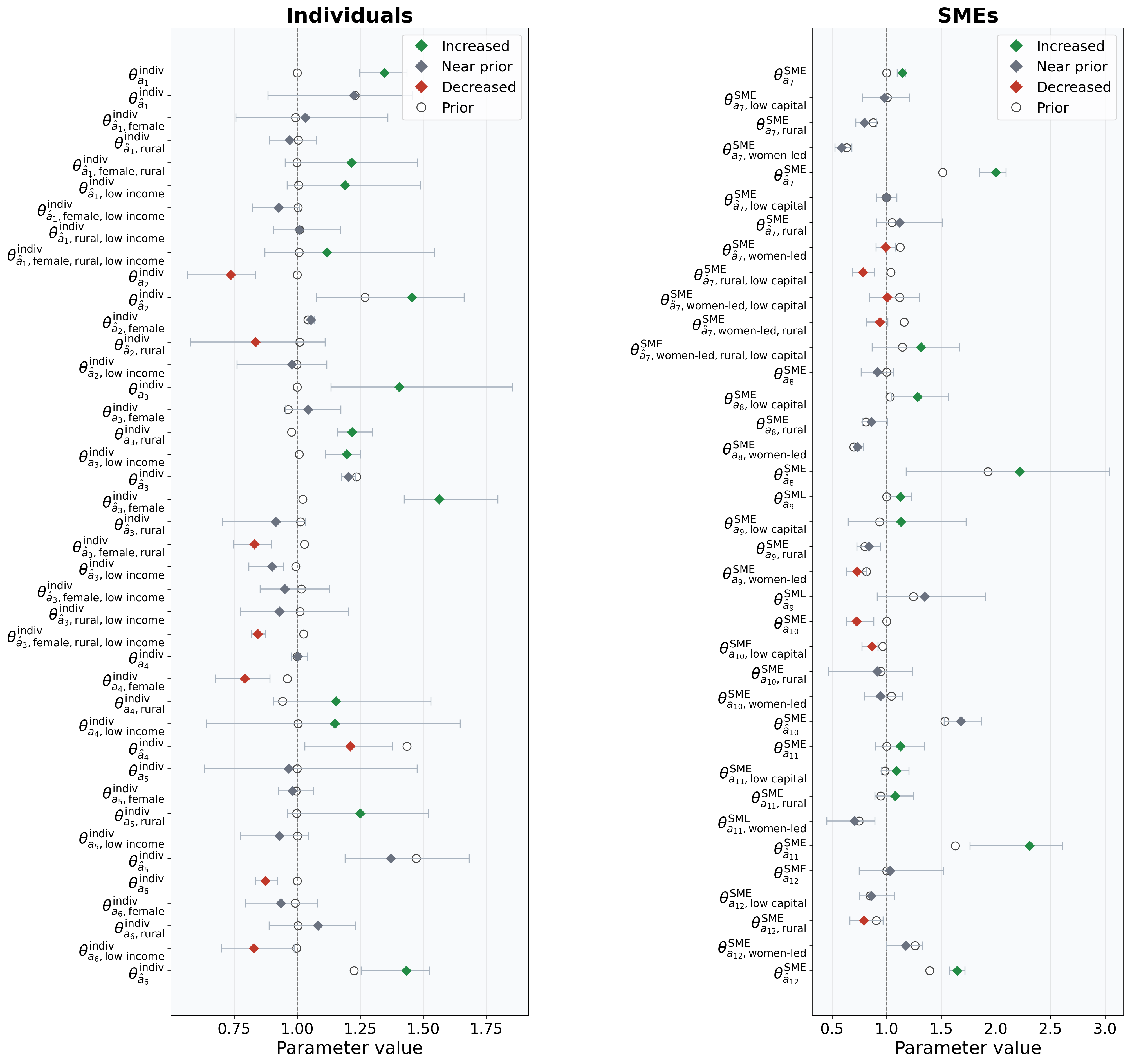}
    \caption{Prior and posterior preference parameters $\theta_j \in \Theta$ after calibration, for individuals (left) and SMEs (right), ordered by action. Circles show the prior mean while diamonds show the posterior mean across the three random seeds, coloured by the direction of the movement (green: increased; red: decreased; grey: within $10\%$ of the prior); horizontal bars show the across-seed range. The dashed line marks the neutral value $\theta_j=1$.}
    \label{fig:calibration}
\end{figure*}

Figure~\ref{fig:calibration} reports the prior and posterior values of the calibrated preference parameters across the three random seeds.
The calibration improves fit to the empirical targets: the weighted root mean square error decreases from $0.101\pm0.001$ under the prior baseline to $0.068\pm0.003$ after calibration, corresponding to a $32.7\%$ reduction in error. The fit is not uniform across agent types: the mean absolute residual is $0.026$ for individual targets but $0.077$ for SME targets, so that SMEs remain harder to reproduce than individuals, and lead to higher standard deviations across seeds. While the aggregate fit is stable across seeds, a number of the subgroup preference terms are only weakly identified: about half of the parameters show a standard deviation above $0.15$, which we flag explicitly in Table~\ref{tab:posterior_largest_movements} and treat as a target for longer chains in the Metropolis-Hastings algorithm and/or additional data in future work.

The calibration results differ a lot across agent types and subgroups, and this is where the hypotheses come from. Because feasibility is kept fixed, every movement in Table~\ref{tab:posterior_largest_movements} is a change in \emph{preference} relative to its prior. Table~\ref{tab:posterior_largest_movements} reports the largest posterior movements, which we turn into a list of hypotheses using a single rule that compares each movement with the group's observed relative uptake of the action. When the two \emph{agree} (preference decreases where uptake is low, or increases where uptake is high) the shortfall is interpreted as a genuine \emph{preference}: the group can act but chooses not to. When they \emph{disagree} (preference increases even though uptake stays low) preference alone cannot explain the gap, so the movement instead is interpreted as unmodelled \emph{opportunity} or \textit{feasibility} constraints: the group would act if it could. Applying this rule to the largest movements provides the subgroup-specific hypotheses H1--H5 below, which the expert panel can accept, revise, or reject during the first validation stage, described in the following validation section\ref{subsec:expertvalidation}.  

\begin{itemize}
\item[\textbf{H1}] \emph{Individuals --- preference.} Women's preference for not using digital payments, $\theta^{\mathrm{indiv}}_{\hat a_3,\mathrm{female}}$, rises above its prior, and women's relative observed use is lower than general rate. \emph{Interpretation:} Low digital gender-gap uptake is consistent with a lower female preference for using digital payments. 
\item[\textbf{H2}] \emph{Individuals --- feasibility.} The rural preferences for digital payments and formal borrowing, $\theta^{\mathrm{indiv}}_{a_3,\mathrm{rural}}$ and $\theta^{\mathrm{indiv}}_{a_5,\mathrm{rural}}$, both increase above their priors, yet rural relative observed uptakes are lower than the general rate. \emph{Interpretation:} These subgroups lower digital-use and formal-borrowing gaps are consistent with lower opportunities, pointing to unmodelled feasibility constraints such as device access, affordability, connectivity, or literacy. The posterior rejects the prior that these groups especially prefer to remain undigitalized.

\item[\textbf{H3}] \emph{SMEs --- preference.} The general complements for not licensing and not hiring, $\theta^{\mathrm{SME}}_{\hat a_7}$ and $\theta^{\mathrm{SME}}_{\hat a_{11}}$, increase, and the digital-sales preference $\theta^{\mathrm{SME}}_{a_{10}}$ decreases, while the observed uptake of all three stays low even though the actions are feasible. \emph{Interpretation:} The general firm's informality and low digital-sales uptake is consistent with lower preferences for formalizing and digitizing.
\item[\textbf{H4}] \emph{SMEs --- feasibility.} The intersectional or subgroup non-license complements of the smallest, rural, and women-led firms fall substantially below their priors. \emph{Interpretation:} These subgroups' lower formalization uptake is consistent with lower opportunities, pointing to unmodelled constraints such as registration cost and administrative complexity, documentation and record-keeping requirements, distance to registration offices, and weak enforcement incentives. The posterior rejects the prior that these groups especially prefer to remain unlicensed.
\item[\textbf{H5}] \emph{Modelling.} The general mobile-money adoption preference $\theta^{\mathrm{indiv}}_{a_2}$ moves sharply downward. \emph{Interpretation:} Unlike H1-H4 this is a movement in the general preference rather than a subgroup deviation, so the agree/disagree contrast (which relies on the subgroup vs. general comparison) does not apply. It can then be interpreted as a correction of a baseline outcome that over-produces wallets relative to the observed share, and is flagged as a modelling issue to revisit rather than a behavioural finding.
\end{itemize}

\begin{table*}[t]
\centering
\caption{Largest posterior movements in percentages relative to prior values, by agent type and averaged across the three random seeds. Each movement is a change in an estimated preference (conditional on the fixed feasibility $\Phi_0$). Movements marked $^{\dagger}$ have a large across-seed spread (posterior s.d. $>0.30$) and are therefore only weakly identified.}
\label{tab:posterior_largest_movements}
\setlength{\tabcolsep}{4pt}
\begin{tabular}{p{1.7cm}p{3.6cm}p{3.4cm}p{1.1cm}p{3.4cm}p{1.1cm}}
\toprule
\multirow{2}{*}{\textbf{Movement}} &
\multirow{2}{*}{\textbf{Interpretation}} &
\multicolumn{2}{c}{\textbf{Individuals}} &
\multicolumn{2}{c}{\textbf{SMEs}} \\
\cmidrule(lr){3-4}\cmidrule(lr){5-6}
& & \textbf{Parameter} & \textbf{Change} &
\textbf{Parameter} & \textbf{Change} \\
\midrule
\multirow{5}{*}{Increased} &
\multirow{5}{3.6cm}{Stronger revealed preference for $a$ (or for remaining in $\hat{a}$) than the prior assumed.}
& $\theta^{\mathrm{indiv}}_{\hat{a}_{3},\mathrm{female}}$ & $+55.7\%$ & $\theta^{\mathrm{SME}}_{\hat{a}_{7}}$ & $+47.8\%$ \\
&& $\theta^{\mathrm{indiv}}_{a_{1}}$ & $+36.5\%$ & $\theta^{\mathrm{SME}}_{\hat{a}_{11}}$ & $+39.9\%^{\dagger}$ \\
&& $\theta^{\mathrm{indiv}}_{a_{3}}$ & $+33.7\%^{\dagger}$ & $\theta^{\mathrm{SME}}_{a_{8},\mathrm{low\ capital}}$ & $+25.7\%^{\dagger}$ \\
&& $\theta^{\mathrm{indiv}}_{a_{5},\mathrm{rural}}$ & $+25.6\%^{\dagger}$ & $\theta^{\mathrm{SME}}_{a_{9},\mathrm{low\ capital}}$ & $+22.6\%^{\dagger}$ \\
&& $\theta^{\mathrm{indiv}}_{a_{3},\mathrm{rural}}$ & $+25.2\%$ & $\theta^{\mathrm{SME}}_{a_{7}}$ & $+18.1\%$ \\
\midrule
\multirow{5}{*}{Decreased} &
\multirow{5}{3.6cm}{Weaker revealed preference for $a$ (or for remaining in $\hat{a}$) than the prior assumed.}
& $\theta^{\mathrm{indiv}}_{a_{2}}$ & $-22.8\%$ & $\theta^{\mathrm{SME}}_{a_{10}}$ & $-27.6\%$ \\
&& $\theta^{\mathrm{indiv}}_{\hat{a}_{3},\mathrm{female,rural,low\ income}}$ & $-22.0\%$ & $\theta^{\mathrm{SME}}_{\hat{a}_{7},\mathrm{women\text{-}led,rural}}$ & $-22.5\%$ \\
&& $\theta^{\mathrm{indiv}}_{\hat{a}_{4}}$ & $-15.5\%$ & $\theta^{\mathrm{SME}}_{\hat{a}_{7},\mathrm{rural,low\ capital}}$ & $-21.7\%$ \\
&& $\theta^{\mathrm{indiv}}_{a_{6},\mathrm{low\ income}}$ & $-15.2\%$ & $\theta^{\mathrm{SME}}_{\hat{a}_{7},\mathrm{women\text{-}led,low\ capital}}$ & $-19.2\%$ \\
&& $\theta^{\mathrm{indiv}}_{a_{4},\mathrm{female}}$ & $-13.4\%$ & $\theta^{\mathrm{SME}}_{\hat{a}_{7},\mathrm{women\text{-}led}}$ & $-17.5\%$ \\
\midrule
\multirow{3}{*}{Near prior} &
\multirow{3}{3.6cm}{Prior already consistent with the data.}
& $\theta^{\mathrm{indiv}}_{a_{4}}$ & $-0.1\%$ & $\theta^{\mathrm{SME}}_{\hat{a}_{8}}$ & $+1.3\%^{\dagger}$ \\
&& $\theta^{\mathrm{indiv}}_{\hat{a}_{5}}$ & $-0.2\%$ & $\theta^{\mathrm{SME}}_{a_{11},\mathrm{low\ capital}}$ & $+2.6\%$ \\
&& $\theta^{\mathrm{indiv}}_{\hat{a}_{1}}$ & $-0.3\%$ & $\theta^{\mathrm{SME}}_{a_{7},\mathrm{rural}}$ & $+2.6\%$ \\
\bottomrule
\end{tabular}
\end{table*}

All in all, for individuals, low digital use has two different hypothetical explanations: an \emph{opportunity} gap for rural groups (H2), and a lower \emph{preference} for women (H1); while the decreasing mobile-money movement is a modelling correction rather than a behavioural finding (H5). For SMEs, high informality and low digitisation are interpreted as a choice for general firms (H3), and as opportunity constraints for the smallest, rural, and women-led firms (H4). SMEs nonetheless remain harder to reproduce than individuals: business licensing is under-predicted at the aggregate ($0.30$ observed versus $0.22$ fitted), and several of the largest SME movements ($\theta^{\mathrm{SME}}_{\hat a_{11}}$, $\theta^{\mathrm{SME}}_{a_9,\mathrm{low\ capital}}$) carry a large across-seed spread and should be treated as weakly identified. Hypotheses H2 and H4 are the ones the framework treats as diagnostic: a revealed preference that moves towards an action whose observed outcome stays low is exactly the signal that the fixed feasibility $\Phi_0$ is missing a constraint for that subgroup, and it is these that the expert panel is asked to confirm or overturn.

Taken together, these results illustrate what the calibration is for. Beyond improving fit, holding feasibility fixed turns each posterior movement into an interpretable statement about behaviour: it differentiates between outcomes that are low because the action is not sufficiently feasible and those that are low because agents, once able, still tend not to act on it. The value of the calibration is therefore the structured, subgroup-specific, and reviewable account of why financial inclusion outcomes could be falling short, considering the gap between outcomes and opportunities, made available to the policy user through this simulation.

\begin{figure}[H]
    \centering
            \begin{minipage}{0.49\textwidth}\centering
        \includegraphics[width=\linewidth]{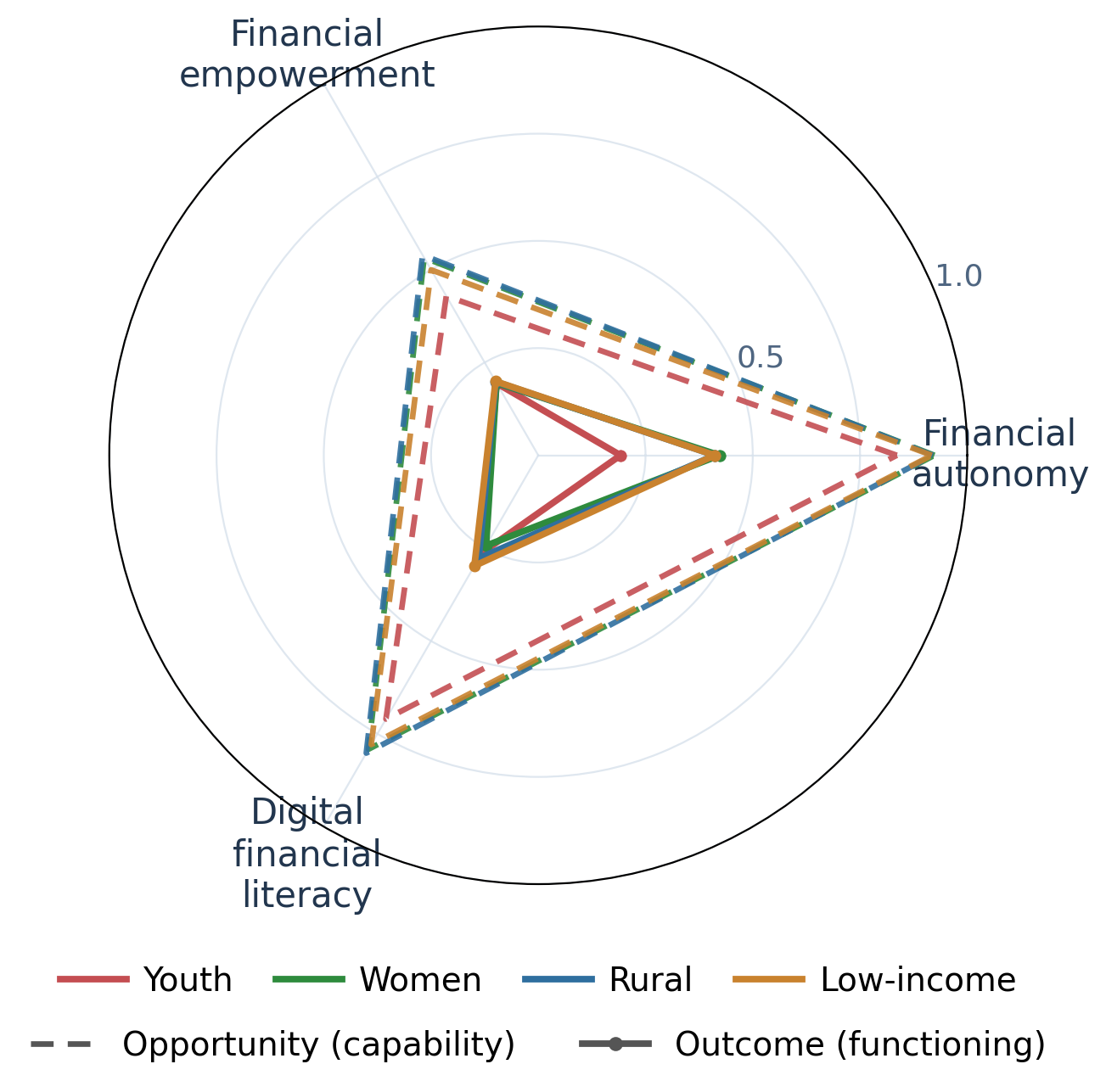}
    \end{minipage}\hfill
    \begin{minipage}{0.49\textwidth}\centering
        \includegraphics[width=\linewidth]{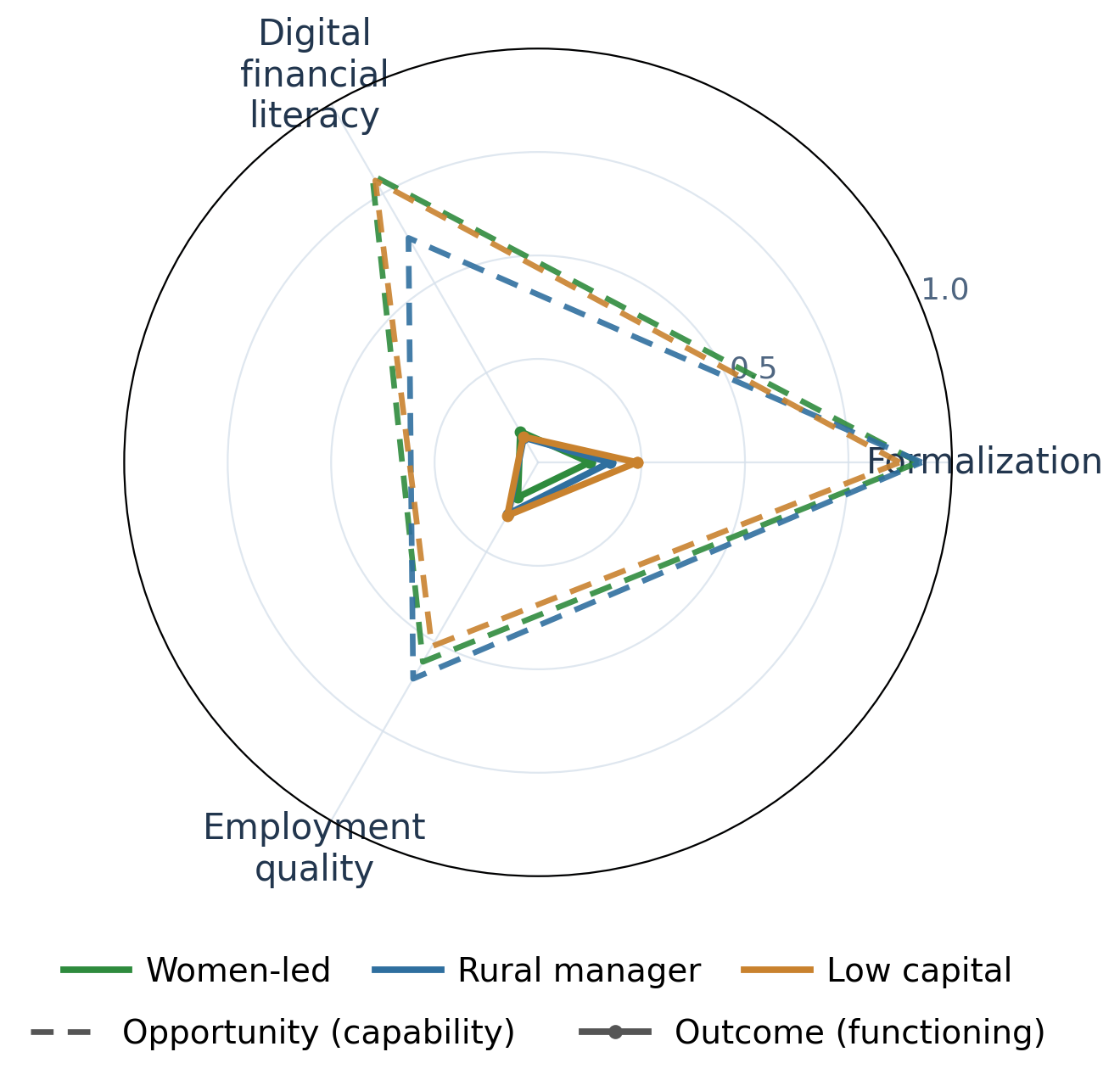}
    \end{minipage}
    \begin{minipage}{0.49\textwidth}\centering
        \includegraphics[width=\linewidth]{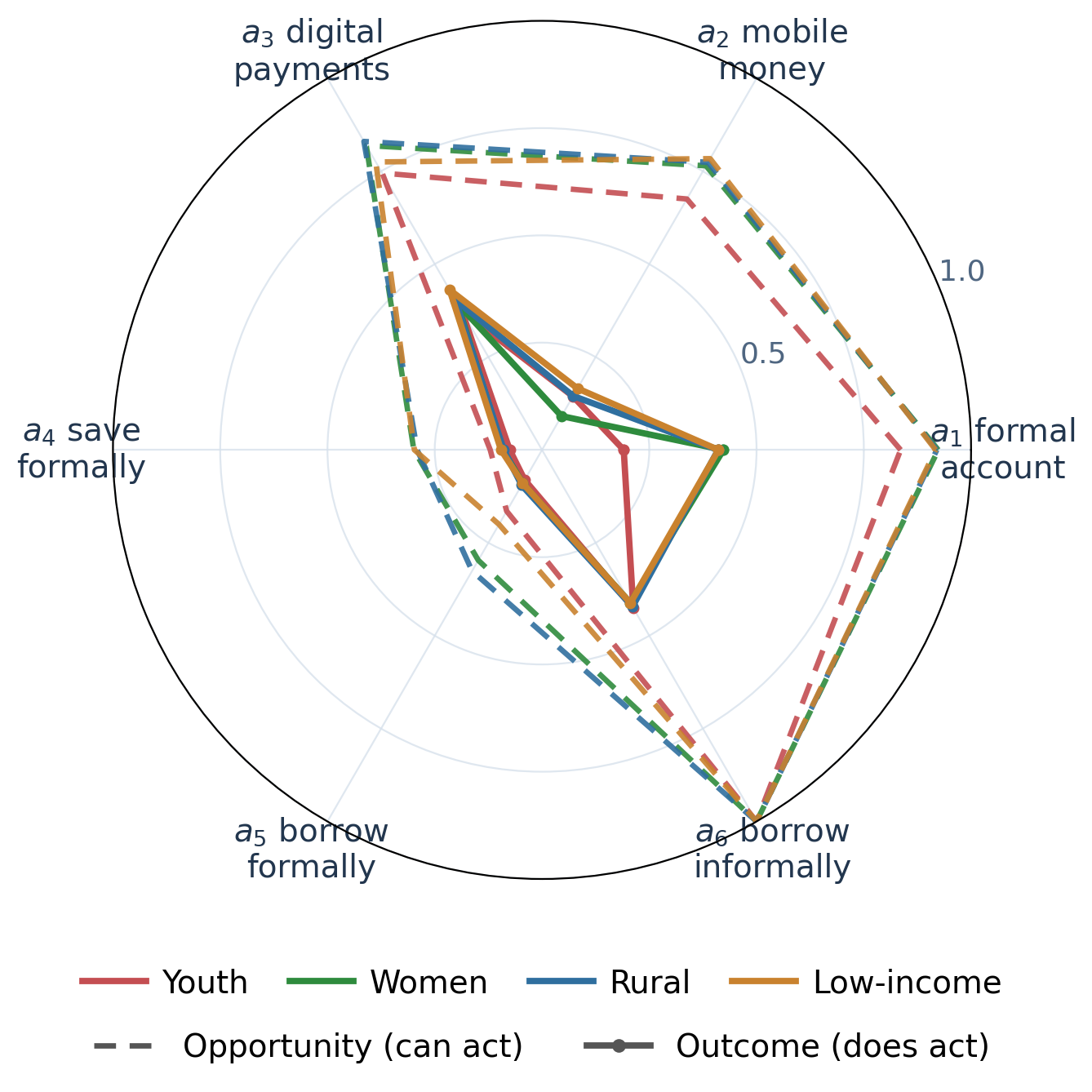}\\
        [-1mm]{\small (a) Individuals}
    \end{minipage}\hfill
    \begin{minipage}{0.49\textwidth}\centering
        \includegraphics[width=\linewidth]{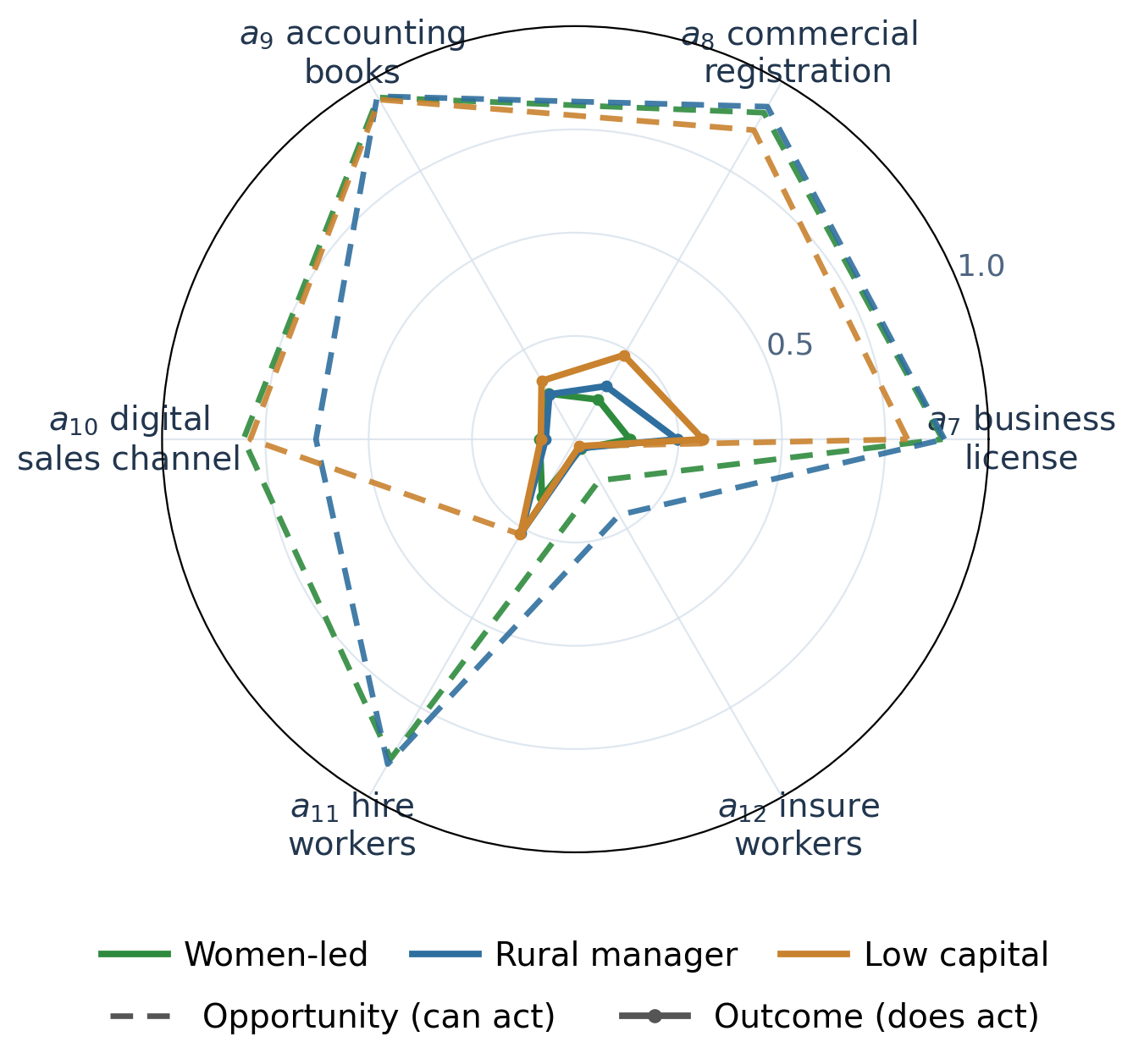}\\
        [-1mm]{\small (b) Small enterprises}
    \end{minipage}
    \caption{Calibrated opportunities (dashed) versus realized outcomes (solid) by target group, for (a) individuals and (b) small enterprises. Colours denote target groups (individuals are red: youth, green: women, blue: rural, amber: low income; and SMEs are green: women-led, blue: rural manager, amber: low capital). For each central capability, its related actions (in Table~\ref{tab:actions}) are aggregated by Eq.~\ref{eq:capability} with equal weights. The distance between the polygons is the capability gap in Eq.~\ref{eq:capabilitygap}. At the action level, the dashed polygon is the share for which the action is a genuine opportunity (feasible above $\tau$ or already realised) and the solid polygon the share that realises it.}   
    \label{fig:capability_outcome}
\end{figure}

Figure~\ref{fig:capability_outcome} makes this opportunity gap explicit at the calibrated state, aggregating the modelled actions into the central financial capabilities and disaggregating each by target group. For every group and capability, the realized outcomes (solid) lie well inside the opportunity set (dashed), revealing an opportunity gap that can be analysed. For individuals, the gap is widest for \emph{financial autonomy} (opportunity around $0.9$ against outcomes near $0.4$) and \emph{digital financial literacy} (opportunity around $0.8$ against outcomes near $0.25$), whereas \emph{financial empowerment} shows both a low frontier and a low outcome (roughly $0.5$ and $0.2$), a sign of structural deprivation rather than an unrealised opportunity. Youth trail the other groups on both opportunity and outcome across all three central capabilities, most starkly on financial autonomy and financial empowerment (outcomes near $0.19$), while rural and low-income individuals show levels close to the population average, consistent with the calibration interpreting their shortfalls as opportunity constraints. For firms the gaps are wider still, reflecting the harder SME fit: \emph{formalization} and \emph{digital financial literacy} show significant gaps for all firm types (opportunities around $0.6$--$0.9$ against outcomes below $0.15$), and low-capital firms sit below the others on \textit{employment} related opportunities, because the model makes hiring barely feasible for the smallest firms. The capability gap is a quantity a policy can target: where the frontier is high but the outcome low, the binding margin is preference; where the frontier itself is low, the binding margin is feasibility. This gap can also be analysed directly at the action level when the action space is small enough (6 actions per agent type in our case), as can be seen in Figure~\ref{fig:capability_outcome}.


\section{Validation}
\label{subsec:expertvalidation}

Even a successfully calibrated model may fail to provide a valid representation of the target system: it may reproduce aggregate targets while misrepresenting individual subgroup patterns, rely on implausible parameter values, or generalize poorly to new sets of data, policies, etc. For this reason, we treat validation as a multi-stage process rather than a single goodness-of-fit exercise, following the classical distinction between building a model with high face validity, validating its assumptions, and comparing its input--output behaviour against the real system~\citep{naylor1967verification}. This paper does not conduct the validation itself. Instead, it defines and motivates the first stage of the proposed validation strategy, which is designed to assess the plausibility of the model's assumptions and calibration outputs through structured domain-expert review.
\begin{figure}[t]
    \centering
    \includegraphics[width=0.9\linewidth]{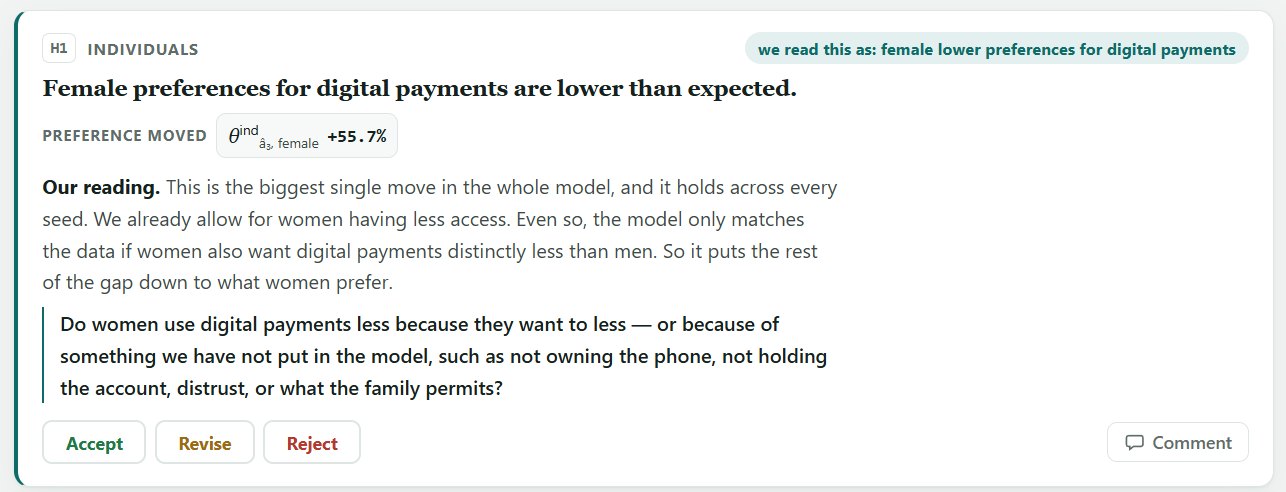}
    \caption{Elicitation item to be shared with the domain-expert panel for hypothesis \textbf{H1}. The expert is shown the parameter movement and our interpretation, and responds accept / revise / reject, with an open prompt to comment on unmodelled barriers that could change it.}
    \label{fig:expert}
\end{figure}
The need for this stage follows directly from the framework. During calibration, the feasibility parameters $\Phi$ are held fixed while the preference parameters $\Theta$ are estimated from observed outcomes. Consequently, a low observed outcome for a particular population subgroup may be explained by low opportunities, low preference for the action, or a combination of both. Calibration therefore produces candidate explanations that remain conditional on the assumed opportunity structure. These are hypotheses rather than validated behavioural conclusions, and require external assessment by experts familiar with the institutional and social context represented by the model. This reflects the equifinality and parameter-identification challenges widely recognised in the empirical validation of agent-based models~\cite{fagiolo2007critical,windrum2007empirical}.

Accordingly, the first validation stage proposed by the framework consists of a structured domain-expert review of both the fixed feasibility assumptions ($\Phi$) and the calibrated preference movements ($\Theta$). Reviewing both components is essential because they jointly determine the interpretation of the calibration results. Accepting a calibrated preference movement also requires accepting the opportunity assumptions against which it was estimated. We here describe the approach for validation with domain experts. Following established guidance on structured expert elicitation~\citep{khodyakov2023rand,sutherland2015policy}, the planned panel is composed of domain experts from central banks, gender- and SME-finance specialists, and Egypt-based field researchers, selected to cover the institutional and social dimensions of the Egyptian context relevant to the model. Each expert receives a summarized description of (i) the simulated agents, (ii) their opportunities and realized actions in the simulation, and (iii) explicit behavioural hypotheses generated from the calibration stage (H1--H5). The first two elements inform the expert about the model, so that they can then provide structured feedback on the third. Figure~\ref{fig:expert} makes this concrete, showing one item as it would be presented to an expert: the movement behind hypothesis H1, with an accept / revise / reject response and an open prompt for missing, unmodelled mechanisms.

\section{Limitations and Ethics}
\label{sec: limitations}
The framework can be applied to more contexts, scaled with more agents, and enriched with more data as it becomes available. However, it will still face some limitations, which lie in the conceptual and numerical assumptions done to reach the results: 
\begin{itemize}
    \item Numerical: We assume numerical baseline and multiplier values for the feasibility definition of actions. We define a numerical prior distribution for preference parameters to be calibrated in Annex~\ref{annex:assumptions} according to observed ratios in the data. For SMEs, additional assumptions are made during synthetic-population construction, since data available for firms is radically different to the individuals one. Each SME is linked to a synthetic manager, and enterprise characteristics are sampled conditional on manager attributes such as sex, age, education, urban/rural status, and workforce status. This preserves broad manager-enterprise relationships, but it does not recover every unobserved correlation between managers, firms, capital, formality, and employment. Also, this makes the SME sample size endogenous rather than fixed. Digital sales are defined using a combined indicator equal to one if the SME has an online page or storefront, uses a mobile sales app, or reports a positive share of online sales; therefore, this variable captures digital market presence rather than direct evidence of digital payment acceptance. Worker variables are reconciled so that any positive non-household-worker count implies hiring, any insured worker implies hiring, and worker-related bands are set to zero for firms without non-household workers. For this reason, worker insurance is measured only among employer SMEs, since firms without non-household workers are not in the relevant risk set.

    \item Conceptual: We assume people's and firms' behaviour can be approximated with the rule-based formula in Eq.~\ref{eq:score}, where actions are prioritized based on feasibility and preference. We assume preferences can be inferred from observed outcomes, although motivations over actions can differ substantially from what people or SMEs end up doing. We also assume calibrated preferences remain fixed over the simulated policy horizon, so policies affect opportunities and feasibility unless they explicitly change preference parameters. Finally, some SME states are interpreted as one-time transitions, such as obtaining a license, registering, hiring non-household workers, or insuring workers. This means the model represents movement from an initial state to a new state, rather than repeated yearly choices over the same outcome.
    
\end{itemize}

As a main ethical consideration, the tool being developed should always be used for probabilistic prescriptions rather than deterministic predictions. Its role is to support decision-making, never to substitute policy-makers. For this reason, the framework emphasizes transparency, interpretability, and human-in-the-loop usage. By explicitly modelling disaggregated opportunities and outcomes through feasibilities and preferences, the approach also provides a tool to surface equity concerns, helping policymakers identify where interventions fail to translate opportunities into meaningful improvements for specific target groups. Additionally, it is important to highlight that the synthetic population and behavioural rules may reflect biases present in the underlying data and modelling assumptions, which can affect groups that are persistently underrepresented, such as people with disabilities. 

\section{Conclusions and Future Work}
\label{sec:conclusions}

This paper presents a general framework for building credible agent-based policy simulations grounded in the Capability Approach and demonstrates its implementation through a case study on financial inclusion in Egypt. The framework integrates four complementary stages: mapping the simulation to the Capability Approach, initialising the model from micro- and macro-level data, calibrating behavioural preferences against observed outcomes, and designing a structured validation strategy that combines empirical evidence with domain-expert knowledge. Together, these stages aim to strengthen the transparency, interpretability, and empirical grounding of policy simulations intended to support real-world decision making.

The financial-inclusion case study illustrates how the framework can be operationalised and already provides policy-relevant insights. By explicitly distinguishing between opportunities and realised outcomes, the Capability Approach enables the calibrated model to identify where financial-inclusion gaps are more likely to arise from limited opportunities and where they are better explained by behavioural preferences. The framework thus transforms calibration from a parameter-fitting exercise into a hypothesis-generation process, producing explicit behavioural hypotheses that can subsequently be assessed by domain experts.

Rather than claiming validated behavioural explanations, we propose the first stage of a broader validation strategy in which the hypotheses generated through calibration are systematically assessed by domain experts. Future work will implement this expert-validation stage and extend the framework through historical input--output validation and benchmarking. Beyond financial inclusion, the proposed methodology can be applied to other policy domains in which the Capability Approach provides an appropriate evaluative framework. More broadly, it also provides the foundation for hybrid decision-support systems in which a validated agent-based model serves as the simulation core of an LLM-assisted policy analysis tool.

\section*{Model and code availability}
In line with the JASSS expectation that models be inspectable, the agent-based model, the synthetic-population generator, and the calibration, evaluation, and figure-generation scripts are available at \url{https://www.comses.net/codebase-release/df8383cb-f49b-4f09-8cce-73603b59adcc/}. The repository includes a structured model description following the ODD protocol~\citep{grimm2020odd}, the frozen structural-baseline configuration ($\Phi_0$) and calibrated preference parameters used for the reported runs, and scripts that reproduce every figure and table in this paper.

\section*{Acknowledgements}
This research has been conducted with the support of the United Nations Economic and Social Commission for Western Asia (UN-ESCWA), which funded this research through the AI-Econ project. We thank the institutional stakeholders, policy practitioners, financial-sector representatives, gender specialists and other domain experts who contributed their time and expertise to the development, contextualization, and validation of the model. 
We also acknowledge the institutions that granted access to the data used to initialize, calibrate, and evaluate the simulation, including ERF and CAPMAS, as well as the collaboration within the EU funded VALAWAI project (101070930) and Spanish funded EVASAI project (PID2024-158227NB-C31), within which part of this research was developed.

\endparano

\clearpage
\newpage
\section{Appendix A: Numerical Assumptions}\label{annex:assumptions}

This appendix documents the numerical values used in the implementation. Table~\ref{tab:feasibility_assumptions} and Table~\ref{tab:feasibility_assumptions2} report the feasibility assumptions that define the fixed set $\Phi$, and Table~\ref{tab:preference_prior_assumptions} reports the prior means of the preference parameters $\Theta$ that enter the calibration.

\begin{table*}[!ht]
\centering
\scriptsize
\caption{Feasibility assumptions, which define action feasibility before preference parameters are applied, classified between baselines, barriers, contextual barriers, and policy multipliers. All complement feasibilities are $1.00$ unless changed by policies, final feasibility is clipped to $[0,1]$ and context barriers qualitative levels are mapped to $0.3$, $0.6$ and $0.9$ quantitative levels. In the 20,000-person synthetic population, about 63\% of individuals report at least one active account-access barrier. For individuals without an active reported barrier, feasibility is informed by the baseline.}
\label{tab:feasibility_assumptions}
\begin{tabular}{p{1cm}p{1.0cm}p{2cm}p{2cm}p{8.2cm}}
\toprule
\textbf{Agent} & \textbf{Action} & \textbf{Parameter ($b_{i,a,t}$ or $b_{a,t}$)} & \textbf{Value / rule} & \textbf{Justification based on domain-expert knowledge} \\

\midrule
\multicolumn{5}{l}{\textit{Baseline feasibility}} \\
\midrule
Individual & $a_1$--$a_6$ & baseline $b^{ind}_{0,a_j}$ & $0.85$; $0.80\,p^\alpha_{i,a_1,t}$; $0.75$; $0.65$; $0.55$; $0.80$ &  Account opening and informal borrowing are considered the most relatively feasible, while usage actions like mobile-money adoption and digital payments are less feasible. The first depends on account-opening feasibility. Formal borrowing is the less feasible action, as hinted by domain-expert reports. \\
SME & $a_7$--$a_{12}$ & baseline $b^{SME}_{0,a_j}$ & $0.70$; $0.65$; $0.60$; $0.65$; $0.55$; $0.50$ & Obtaining a business license is the most feasible SME transition, followed by commercial registration and adopting a digital sales channel, then accounting books. Hiring non-household workers and, especially, insuring workers receive the lowest baselines; worker insurance is the lowest because it requires both formal employment relationships and a connection to labour or social-insurance institutions.  \\ 

\midrule
\multicolumn{5}{l}{\textit{Barrier multipliers}} \\
\midrule
Individual & $a_1$, $a_4$ & not enough money & $\times0.20$, $\times0.30$ & Lack of money strongly reduces opening an account and saving formally. Domain-expert reports mark insufficient funds as the most frequent unbanked barrier in Arab LMICs, with Egypt about 90\% and Morocco about 75\% among the unbanked. \\ 
& $a_1$ & high fees, strict documentation & $\times0.05$, $\times0.10$ &  High fees and hard documentation requirements make account opening difficult. Global Findex and ESCWA Financial Inclusion Index evidence identify account costs and documentation requirements among the most commonly reported barriers to account ownership in developing economies and Arab countries, although their relative importance is generally lower than insufficient funds \citep{KlapperEtAl2025,ESCWA2025FII}. \\

& $a_1$, $a_3$ & distance & $\times0.50$, $\times0.70$ & Distance is treated as a stronger constraint for opening a formal account than for digital payments. Account opening often requires reaching a branch, agent, or formal provider at least once, while digital payments can sometimes be performed remotely once a usable channel exists. Digital use barriers are mainly because of weaker agent networks, weaker connectivity, and poorer service density. This is especially relevant for rural populations, with Morocco reporting a clear rural/urban gap in the ESCWA/FII evidence. \\
& $a_3$, $a_4$, $a_5$ & low self-efficacy & $\times0.50$, $\times0.70$, $\times0.70$ & Low self-efficacy captures limited ability to navigate providers or products, which reduces digital use, saving, and formal borrowing. The penalty is strongest for digital payments because digital payment use requires active operation of a device, interface, PIN, wallet, or application. This is consistent with Global Findex evidence showing that account ownership does not automatically translate into active use of digital payments, savings, or credit products, and with GSMA Mobile Gender Gap reports identifying digital skills, literacy, confidence, and safety concerns as major barriers to mobile internet and digital financial service adoption, particularly among women and low-income populations \citep{KlapperEtAl2025,GSMA2026MobileGenderGap}. \\

& $a_1$, $a_3$, $a_4$, $a_5$ & income level, ~~~~~work status & low: $a_1:\times0.85$, $a_4:\times0.75$, $a_5:\times0.75$; high: $a_1,a_3,a_4:\times1.10$; not working: $a_4:\times0.85$, $a_5:\times0.70$ &
Income level is treated as a structural determinant of financial inclusion rather than a direct barrier. Low income modestly reduces account opening and more strongly lowers formal saving and borrowing because these actions require disposable income and repayment capacity. Higher income slightly increases account opening, digital use, and formal saving by easing affordability and access constraints. Effects are kept moderate because income alone does not remove barriers such as trust, distance, or product design. This is consistent with Global Findex evidence showing higher rates of account ownership, digital payments, and formal saving among higher-income adults, while financial exclusion remains concentrated among poorer populations. Being outside the workforce is treated as a structural constraint on formal saving and formal borrowing. It lowers expected income regularity and repayment capacity, so the reduction is stronger for formal borrowing than for saving.\\
\midrule
\multicolumn{5}{l}{\textit{Continued on next page}} \\

\bottomrule
\end{tabular}

\begin{minipage}{0.98\linewidth}
\vspace{0.4em}
\scriptsize
\textit{Reading the table. } Start with the action baseline, then multiply by every applicable observed barrier or structural condition, context term, and pre-calibration baseline scale. In the reference no-policy model, multiply by $\alpha=1$ and clip the resulting feasibility to $[0,1]$. 

\end{minipage}
\end{table*}

\begin{table}
\scriptsize
\caption{Remaining barrier multipliers for SMEs and context barrier multipliers.}
\label{tab:feasibility_assumptions2}
\begin{tabular}{p{1cm}p{1.0cm}p{1.6cm}p{3cm}p{8cm}}
\toprule
\textbf{Agent} & \textbf{Action} & \textbf{Parameter} & \textbf{Value / rule} & \textbf{Justification based on domain-expert knowledge} \\

\midrule
\multicolumn{5}{l}{\textit{Barrier multipliers}} \\
\midrule
SME & $a_7$--$a_9$ & formalization readiness score $r_f$ & base $0.55$; capital low/mid/high: $\times0.45$, $\times1.00$, $\times1.35$; workers no/yes: $\times0.75$, $\times1.35$; age young/mid/older: $\times0.85$, $\times1.00$, $\times1.10$; manager income low/mid/high: $\times0.85$, $\times1.00$, $\times1.10$; $r_f$ clipped to $[0.15,1.50]$; $a_7,a_8:\times r_f$; $a_9:\times(0.65+0.35r_f)$ &
Data-informed structural proxy. Since barriers are not explicitely reported in data, we create a formalization readiness score. The score starts at $0.55$ and is adjusted using observed SME characteristics.
Small, young, low-capital firms without workers are treated as less ready to formalize.
Larger, older, higher-capital firms with non-household workers are treated as more ready.
License and registration depend directly on $r_f$; accounting books depend on it more softly because bookkeeping can occur before full formalization.
The direction is supported by ELMPS SME variables and MSME-barrier reports; exact multipliers are modelling assumptions. \\
SME 
& $a_7$--$a_{9}$ & manager has formal finance & $\times1.40$, $\times1.50$, $\times1.20$ & Prior use of formal finance increases familiarity with documentation, record keeping, and provider requirements, which makes licensing, registration and bookkeeping more feasible. The penalty for managers without formal-finance experience captures exclusion among small and informal firms (World Bank Enterprise Surveys, Egypt 2020; Morocco 2019). \\
SME & $a_{10}$, $a_9$ & manager digital state & $a_{10}:\times1.80$, $a_9:\times1.20$; plus $r_f:\times1.05$, followed by clipping $r_f$ to $[0.15,1.50]$ & Digital capability is a key determinant of SME adoption of digital payments and sales channels. World Bank and GSMA evidence shows that adoption depends not only on infrastructure but also on managerial skills, confidence, and ability to use digital tools. The bookkeeping effect is smaller because records can still be maintained manually. The readiness bonus reflects complementarities between digitalization and formalization. \\
& $a_{10}$,$a_7$ & rural manager & $a_{10}:\times0.80$, $a_7:\times0.90$ & Rural firms face weaker connectivity, lower digital adoption among customers, and fewer nearby service providers. The stronger penalty is therefore assigned to digitization. Licensing is less affected because administrative procedures can often be completed through intermediaries or occasional travel, although distance and information costs remain relevant. This is consistent with rural–urban gaps reported in ESCWA financial inclusion and GSMA digital-access evidence. \\
& $a_{11}$, $a_{12}$ & capital scale & low capital: $a_{11}:\times0.60$, $a_{12}:\times0.50$; high capital: $a_{11}:\times1.50$, $a_{12}:\times1.60$  & Capital is one of the strongest predictors of firm growth and formal employment. Hiring and worker insurance require sustained wage and compliance costs, while productive investment can sometimes be financed incrementally through retained earnings or supplier credit. Enterprise Survey data consistently show higher investment, employment growth, and formal-finance access among larger and better-capitalized firms, justifying the stronger effects on hiring and insurance. \\
& $a_7$, $a_8$ & older enterprise & $\times1.15$, $\times1.15$ & Older firms are generally more stable, visible, and likely to have accumulated administrative knowledge and business records. Enterprise Survey evidence shows that older firms are more likely to be formal and to interact with government and financial institutions. This effect is kept modest because firm age is already partially captured in the readiness score $r_f$. \\
& $a_{12}$, $a_9$, $a_8$ & missing formal state & no license and no registration: $\times0.50$, $\times0.55$, $\times1.00$; no license: $\times1.00$, $\times1.00$, $\times0.70$ & Worker insurance typically requires formal registration and interaction with labor or social-insurance systems, making it difficult for fully informal firms. Bookkeeping is easier but still less common among informal enterprises. Registration itself is not penalized by lacking registration because it is the target transition; however, licensing may facilitate registration where administrative procedures are sequential. This ordering reflects standard formalization pathways documented in World Bank SME and Enterprise Survey evidence. \\
& $a_{11}$, $a_{12}$ & existing state at choice time & already hires: $a_{11}:\times0.15$; no hiring yet: $a_{12}=0$; already insured: $a_{12}=0$ & Repeating hiring or investment is slowed; worker insurance is much less feasible before hiring workers. \\

\midrule
\multicolumn{5}{l}{\textit{Context barrier multipliers}} \\
\midrule
Individual & $a_1$ & account fees context & $\times(1-0.30\,\beta_f)$ & Higher contextual fees lower account-opening feasibility. \\
& $a_1$ & documentation context & $\times(1-0.25\,\beta_d)$ & Stricter documentation lowers account-opening feasibility. \\
& $a_1$ & traditional infrastructure & $\times(0.50+0.50\,\beta_I)$ & Better physical financial infrastructure raises account-opening feasibility. The $0.50$ floor means access is reduced but not automatically zero when infrastructure is weak. \\
& $a_4$, $a_5$ & traditional infrastructure & $\times(0.60+0.40\,\beta_I)$ & Better physical infrastructure supports formal saving and formal borrowing, with a $0.60$ floor because these actions also depend on agent state and barriers. \\
& $a_3$ & digital infrastructure & $\times \beta_D$ & Digital infrastructure directly scales digital-payment feasibility. \\
& $a_5$ & formal trust context & $\times \beta_T$ & Higher formal trust raises formal borrowing feasibility. \\
SME & $a_7$, $a_8$ & documentation context & $\times(1-0.35\,\beta_d)$ & Documentation is treated as a stronger contextual barrier for firm formalization than for individual account opening. \\
& $a_9$, $a_{11}$ & traditional infrastructure & $\times(0.70+0.30\,\beta_I)$ & Infrastructure supports accounting and hiring, but these actions retain a high floor because firm characteristics also matter. \\
& $a_{12}$ & traditional infrastructure & $\times(0.60+0.40\,\beta_I)$ & Worker insurance is more infrastructure-dependent than bookkeeping or hiring. \\
& $a_{10}$ & digital infrastructure & $\times \beta_D$ & Digital infrastructure directly scales SME digitization. \\

\midrule

\end{tabular}
\end{table}

\begin{table*}[!tp]
\centering
\scriptsize
\caption{Preference-prior assumptions classified between neutral or data-informed. Prior mean values $\theta^0$ used in the simulation are listed by agent type, action set, and associated mathematical parameter. A justification explicitely explain how are they built from observed shrinked ratios $R$ in the data, i.e. 20,000-person synthetic population and the associated 4,318-row synthetic SME. Additional shrinkage is considered for extremely large ratios, only when $\sqrt{R}>2$. }
\label{tab:preference_prior_assumptions}
\begin{tabular}{p{1cm}p{1.2cm}p{2.1cm}p{2.0cm}p{7.9cm}}
\toprule
\textbf{Agent} & \textbf{Action} & \textbf{Parameter(s)} & \textbf{Value ($\sqrt R \rightarrow \theta^0$)} & \textbf{Justification based on observed ratios} \\
\midrule
\multicolumn{5}{l}{\textit{Neutral or weakly identified preference priors}} \\
\midrule
Individual &
$a_1$--$a_6$ &
$\theta^{\mathrm{indiv}}_{a_j}$ &
$1.00$ &
Neutral prior for overall individual transition actions. Calibration can move the values up or down using synthetic-person targets, ratios are considered in the complement actions. \\

SME &
$a_7$--$a_{12}$ &
$\theta^{\mathrm{SME}}_{a_j}$ &
$1.00$ &
Neutral prior for overall SME transition actions. Calibration can move the values up or down using synthetic-SME targets when available. \\


\midrule
\multicolumn{5}{l}{\textit{Data-informed preference priors below one}} \\
\midrule
Individual &
$a_3$  &
$\theta^{\mathrm{indiv}}_{a_3,\mathrm{female}}$, $\theta^{\mathrm{indiv}}_{a_3,\mathrm{rural}}$ &
$0.965$, $0.978$ &
Data-informed priors for digital-use transition actions, with ratios for female and rural subgroups equal to $0.368/0.396=0.931 \rightarrow \theta^0=0.965$ and $0.378/0.396=0.956 \rightarrow \theta^0=0.978$, respectively.  \\ 

Individual &
$\hat a_3$  &
$\theta^{\mathrm{indiv}}_{\hat a_3,\mathrm{low\ income}}$ &
$0.995$ &
Data-informed prior for non-digital use complement action, with ratio for low-income group equal to $0.598/0.604=0.989 \rightarrow \theta^0=0.995$. \\

Individual &
$a_4$  &
$\theta^{\mathrm{indiv}}_{a_4,\mathrm{female}}$ &
$0.961$ &
Data-informed prior for formal-saving transition action, with ratio for female subgroup equal to $0.083/0.090=0.924 \rightarrow \theta^0=0.961$. \\

SME &
$a_7$--$a_9$, $a_{11}$ profiles &
$\theta^{\mathrm{SME}}_{a_j,g}$ &
$0.633$--$0.876$ &
Data-informed priors for SME transition actions with ratios for women-led license, rural-manager license, women-led registration, women-led accounting books, and women-led hiring equal to $0.122/0.304=0.401 \rightarrow \theta^0=0.633$, $0.233/0.304=0.767 \rightarrow \theta^0=0.876$, $0.103/0.212=0.488 \rightarrow \theta^0=0.698$, $0.111/0.168=0.662 \rightarrow \theta^0=0.814$, and $0.1533/0.2740=0.5595 \rightarrow \theta^0=0.748$, respectively.  \\

SME &
$\hat a_7$  &
$\theta^{\mathrm{SME}}_{\hat a_7,\mathrm{low\ capital}}$ &
$0.998$ &
Data-informed prior for the license non-adoption complement action, with ratio for low-capital subgroup equal to $0.693/0.696=0.996 \rightarrow \theta^0=0.998$.  \\

\midrule
\multicolumn{5}{l}{\textit{Data-informed preference priors above one}} \\
\midrule
Individual &
$\hat a_1$--$\hat a_6$ &
$\theta^{\mathrm{indiv}}_{\hat a_j}$ &
$1.083$--$1.226$ &
Data-informed priors for individual complement actions, using ratios equal to: no account $0.540/0.460=1.173 \rightarrow \theta^0=1.083$; no mobile money $0.846/0.154=5.500$, with $\sqrt R=2.345 \rightarrow \theta^0=1.269$; no digital payments $0.604/0.396=1.527 \rightarrow \theta^0=1.236$; no formal saving $0.910/0.090=10.093$, with $\sqrt R=3.177 \rightarrow \theta^0=1.435$; no formal borrowing $0.919/0.081=11.293$, with $\sqrt R=3.360 \rightarrow \theta^0=1.472$; no informal borrowing $0.601/0.399=1.504 \rightarrow \theta^0=1.226$. Large ratios are further shrunk after $\sqrt R$.\\

Individual &
$\hat a_1$  &
$\theta^{\mathrm{indiv}}_{\hat a_1,g}$ &
$1.002$--$1.009$ &
Data-informed priors for non-account complement actions, using ratios for:  female $0.542/0.540=1.005 \rightarrow \theta^0=1.002$; rural $0.546/0.540=1.012 \rightarrow \theta^0=1.006$; low income $0.542/0.540=1.004 \rightarrow \theta^0=1.002$; female-rural $0.548/0.540=1.015 \rightarrow \theta^0=1.007$; female-low-income $0.542/0.540=1.004 \rightarrow \theta^0=1.002$; rural-low-income $0.549/0.540=1.018 \rightarrow \theta^0=1.009$; female-rural-low-income $0.545/0.540=1.009 \rightarrow \theta^0=1.005$.\\

Individual &
$\hat a_3$  &
$\theta^{\mathrm{indiv}}_{\hat a_3,g}$ &
$1.012$--$1.030$ &
Data-informed priors for non-digital use complement actions, using ratios: female $0.632/0.604=1.045 \rightarrow \theta^0=1.022$; rural $0.622/0.604=1.029 \rightarrow \theta^0=1.014$; female-rural $0.641/0.604=1.060 \rightarrow \theta^0=1.030$; female-low-income $0.625/0.604=1.034 \rightarrow \theta^0=1.017$; rural-low-income $0.619/0.604=1.024 \rightarrow \theta^0=1.012$; female-rural-low-income $0.637/0.604=1.054 \rightarrow \theta^0=1.027$. \\

Individual &
$a_3$, $a_5$  &
$\theta^{\mathrm{indiv}}_{a_3,\mathrm{low\ income}}$, $\theta^{\mathrm{indiv}}_{a_5,\mathrm{low\ income}}$ &
$1.008$, $1.001$ &
Data-informed priors for transition actions, using low-income ratios of digital use $0.402/0.396=1.016 \rightarrow \theta^0=1.008$, and low-income formal borrowing $0.065/0.064=1.003 \rightarrow \theta^0=1.001$. \\

SME &
$\hat a_7$--$\hat a_{12}$ &
$\theta^{\mathrm{SME}}_{\hat a_j}$ &
$1.245$--$1.929$ &
Data-informed priors for complement actions, using ratios for no license $0.696/0.304=2.291$; no registration $0.788/0.212=3.719$; no books $0.832/0.168=4.948$; no combined digital sales channel $0.931/0.069=13.490$; no reconciled non-household hiring $0.7260/0.2740=2.650$; and uninsured/insured among employer SMEs $0.8986/0.1014=8.858$. \\

SME &
$\hat a_7$   &
$\theta^{\mathrm{SME}}_{\hat a_7,g}$ &
$1.038$--$1.160$ &
Data-informed priors for non-license adoption complement actions, using subgroup ratios: women-led $0.878/0.696=1.262 \rightarrow \theta^0=1.123$; rural-manager $0.767/0.696=1.102 \rightarrow \theta^0=1.050$; women-led-rural $0.936/0.696=1.345 \rightarrow \theta^0=1.160$; women-led-low-capital $0.872/0.696=1.252 \rightarrow \theta^0=1.119$; rural-low-capital $0.750/0.696=1.077 \rightarrow \theta^0=1.038$; women-led-rural-low-capital $0.913/0.696=1.312 \rightarrow \theta^0=1.145$. \\
 \\

SME &
$a_7$ &
$\theta^{\mathrm{SME}}_{a_7,\mathrm{low\ capital}}$ &
$1.005$ &
Data-informed prior for the license-adoption transition action, using ratio for low capital subgroup equal to $0.307/0.304=1.010 \rightarrow \theta^0=1.005$. \\

\bottomrule
\end{tabular}

\begin{minipage}{0.98\linewidth}
\vspace{0.4em}
\scriptsize
\textit{Reading the table.} A value of $1$ means no prior direction. A value below $1$ gives the calibration a weak starting tendency below neutral; a value above $1$ gives it a weak starting tendency above neutral. These are starting values only, not final calibrated results.

\end{minipage}
\end{table*}

\bibliographystyle{jasss}
\bibliography{references} 


\end{document}